\documentclass{article}

\usepackage{microtype}
\usepackage{graphicx}
\usepackage{booktabs} %

\usepackage{hyperref}
\usepackage{subfigure}

\usepackage[accepted]{arxiv_icml2025}

\usepackage{amsmath}
\usepackage{amssymb}
\usepackage{mathtools}
\usepackage{amsthm}

\usepackage[capitalize,noabbrev]{cleveref}

\theoremstyle{plain}

\theoremstyle{definition}

\theoremstyle{remark}

\usepackage[textsize=tiny]{todonotes}

\usepackage{amsmath,amsfonts,bm}

\def\secref#1{section~\ref{#1}}
\def\Secref#1{Section~\ref{#1}}

\def\eqref#1{equation~\ref{#1}}

\def\1{\bm{1}}

\DeclareMathAlphabet{\mathsfit}{\encodingdefault}{\sfdefault}{m}{sl}
\SetMathAlphabet{\mathsfit}{bold}{\encodingdefault}{\sfdefault}{bx}{n}

\usepackage{url}
\usepackage{dsfont}
\usepackage{stfloats}

\usepackage{booktabs}
\usepackage{makecell}
\usepackage{hyperref}
\usepackage{amsmath}
\hypersetup{
  colorlinks=true,
  urlcolor=blue
}

\usepackage{comment}
\usepackage{multirow,bigdelim}
\usepackage{lipsum}
\usepackage{array}
\usepackage{wrapfig}

\usepackage[percent]{overpic}
\usepackage{makecell}
\usepackage{blindtext}
\usepackage{xcolor}
\usepackage{soul}
\usepackage{cleveref}
\usepackage{amsfonts}
\usepackage{xspace} %
\usepackage{braket}
\usepackage{nicefrac}
\usepackage{enumitem} %
\usepackage{multicol} %
\usepackage{dsfont}
\usepackage[normalem]{ulem}
\usepackage{soul}
\usepackage{wrapfig}

\makeatletter
\newcommand{\settitle}{\@maketitle}
\makeatother

\newcolumntype{C}[1]{>{\centering\let\newline\\\arraybackslash\hspace{0pt}}m{#1}}

\newif\ifdraft
\drafttrue

\newcommand{\tildeapprox}{{\raise.17ex\hbox{$\scriptstyle\sim$}}}
\makeatletter
\@ifundefined{secref}
  {\newcommand{\secref}[1]{\Secref{#1}}}
  {\renewcommand{\secref}[1]{\Secref{#1}}}
\makeatother

\makeatletter
\@ifundefined{figref}
  {}
  {}
\makeatother

\renewcommand{\cref}[1]{Fig. \ref{#1}}

\renewcommand{\eqref}[1]{Eq.~(\ref{#1})}

\definecolor{darkpink}{rgb}{0.561, 0.282, 0.427}
\definecolor{atomictangerine}{rgb}{0.8, 0.2, 0.1}
\definecolor{turq}{rgb}{0.0, 0.5, 0.5}
\definecolor{darkturq}{rgb}{0.0, 0.4, 0.4}
\definecolor{bright}{rgb}{0.8, 0.1, 0}
\definecolor{darkgray}{gray}{0.3}
\definecolor{gray}{gray}{0.5}
\definecolor{mahogany}{rgb}{0.6, 0.05, 0.05}
\definecolor{editblue}{rgb}{0.3,0.05,0.9}
\definecolor{black}{rgb}{0.,0.,0.}
\definecolor{darkgreen}{rgb}{0.1,0.5,0.0}
\definecolor{olive}{rgb}{0.537, 0.627, 0.318}
\definecolor{green}{rgb}{0.22, 0.463, 0.114}
\definecolor{grey}{rgb}{0.4, 0.4, 0.4}
\definecolor{blue}{rgb}{0.435, 0.659, 0.863}
\definecolor{pink}{rgb}{0.761, 0.482, 0.627}
\definecolor{darkpink}{rgb}{0.561, 0.282, 0.427}

\ifdraft

\definecolor{myblue}{rgb}{0.1,0.05,0.8}

\newcommand\edit[1]{\textcolor{black}{ #1 \!\!}}
\makeatletter
\@ifundefined{todo}
  {}
  {}
\makeatother

\newcommand\yoni[1]{}

\newcommand\gal[1]{}
\newcommand\yuval[1]{}
\newcommand\yoad[1]{}

\newcommand{\drop}[1]{}

\else
\newcommand{\dcc}[1]{}
\newcommand{\rgc}[1]{}
\newcommand{\opc}[1]{}
\newcommand{\gcc}[1]{}
\newcommand{\hmc}[1]{}
\newcommand{\abc}[1]{}

\newcommand\gal[1]{}
\newcommand\yuval[1]{}
\newcommand\yoad[1]{}
\fi

\newcommand{\ourmethod}{\textit{Video Storyboarding}\xspace}

\newcommand{\TTV}{T2V\xspace}

\makeatletter
\DeclareRobustCommand\onedot{\futurelet\@let@token\@onedot}
\def\@onedot{\ifx\@let@token.\else.\null\fi\xspace}

\def\eg{\emph{e.g}\onedot}

\makeatother
\usepackage{sidecap}
\usepackage{graphicx}

\usepackage[bottom]{footmisc}
\usepackage{arydshln}

\makeatletter
\def\blfootnote{\xdef\@thefnmark{}\@footnotetext}
\makeatother

\usepackage{colortbl} %

\begin{document}

\twocolumn[
\icmltitle{Motion by Queries: Identity-Motion Trade-offs in Text-to-Video Generation.
}

\begin{icmlauthorlist}
\icmlauthor{Yuval Atzmon}{yyy}
\icmlauthor{Rinon Gal}{yyy}
\icmlauthor{Yoad Tewel}{yyy}
\icmlauthor{Yoni Kasten}{yyy}
\icmlauthor{Gal Chechik}{yyy}
\end{icmlauthorlist}

\icmlaffiliation{yyy}{NVIDIA}

\icmlcorrespondingauthor{Yuval Atzmon}{yatzmon@nvidia.com}

\icmlkeywords{Machine Learning, ICML}

\vskip 0.3in
]

\printAffiliationsAndNotice{} %

\begin{abstract}
Text-to-video diffusion models have shown remarkable progress in generating coherent video clips from textual descriptions. However, the interplay between motion, structure, and identity representations in these models remains under-explored. Here, we investigate how self-attention query ($Q$) features simultaneously govern motion, structure, and identity and examine the challenges arising when these representations interact. Our analysis reveals that Q affects not only layout, but that during denoising Q also has a strong effect on subject identity, making it hard to transfer motion without the side-effect of transferring identity. Understanding this dual role enabled us to control query feature injection (Q injection) and demonstrate two applications: (1) a zero-shot motion transfer method -- implemented with VideoCrafter2 and WAN 2.1 -- that is 10$\times$ more efficient than existing approaches, and (2) a training-free technique for consistent multi-shot video generation, where characters maintain identity across multiple video shots while Q injection enhances motion fidelity.

\end{abstract}

\section{Introduction}

Video generation from text is at the forefront of generative AI. Despite progress in controlling entities in video, several major challenges remain, including generating natural, engaging motion and preserving consistent identity of entities throughout the video. These two goals often form a trade-off: It is easy to preserve consistency if motion is strongly limited, and making entities move makes it harder to enforce consistency, as the appearance of an entity changes.  
A major challenge lies in understanding how motion and identity are represented in various components of video generation models and how to effectively control them.

This limited understanding hinders downstream applications in video generation. 
For instance, while many current motion transfer approaches \cite{yatim2024space, zhao2024motiondirector, motion_inversion, lin2025equivdm} rely on tuning or test-time optimization, there is growing interest in developing inference-time methods, similar to how text-to-image layout transfer already operates through feature manipulation~\cite{cao_2023_masactrl,alaluf2023crossimage}. Better model understanding could lead to further progress in this direction. As another example, consider consistent characters in multi-shot video generation, where the goal is to preserve consistency of character identity and appearance across shots. Image-based models tackle this through feature-sharing, but applying the same ideas to video leads to loss of motion because the shared features encode both identity \textit{and} motion.

\begin{figure}[htbp]
    \centering

    \includegraphics[width=0.75\linewidth,trim={0.cm 21.2cm 23cm 0cm},clip] {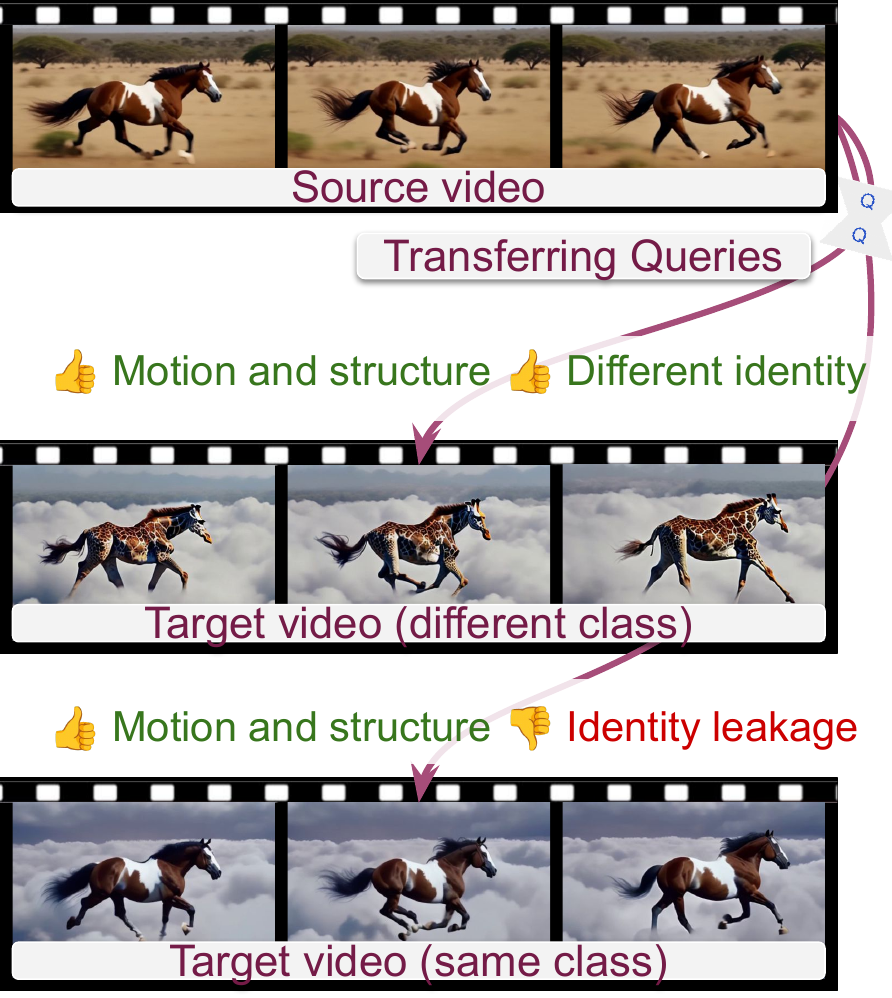} %
    \caption{ \textbf{\href{https://research.nvidia.com/labs/par/MotionByQueries/\#teaser}{(click-to-view-online)}} Our analysis reveals differences in Q-injection between text-to-video and text-to-image models. One key observation is that in text-to-video models, zero-shot Q injection can transfer structure and motion between different video shots. However, when a target video is prompted for the same subject, Q injection suffers from identity leakage.}
        \label{fig1}

\end{figure}

To understand representation of motion and identity in a video model, let us first draw analogies from text-to-image (T2I) models which are better understood. Motion can be viewed as a "3D-shape" in the 3D tensor defined by a sequence of frames, so it is natural to look at representation of shape and structure in image generation models.
Previous works showed that diffusion-based T2I models establish the layout of the generated image in early steps~\citep{patashnik2023localizing}. Several studies~\cite{cao_2023_masactrl,alaluf2023crossimage,tewel2024training} analyzed how different model components determine this structure. They showed that self-attention queries (Q) encode structural layout information, and injecting queries between images during generation ``copies" shape while preserving appearance. As motion is the video equivalent of structure, we investigate the relationship between query vectors, motion, and identity in video generation.

We conduct an empirical analysis, and observe that in contrast to image models, Q in video-generation models affects both motion and identity. Moreover, videos require more denoising steps than images to capture motion patterns. We use this insight to control motion in two different applications: motion transfer and consistent multi-shot video generation.

For motion transfer, we find that injecting Q features from a source video during denoising of a generated video allows transferring motion to a new video in a zero-shot manner (without extra fine-tuning or optimization).
Our simple pipeline achieves generation quality close to leading methods, while being 10$\times$ more efficient than existing approaches.

For consistent multi-shot video generation, we build on insights from multi-shot image generation \cite{tewel2024training} relying on shot-to-shot extended attention. We find that Q injection from unconstrained video generation can preserve layout and motion diversity. However, unlike images, video generation requires more Q injection steps, causing identity leakage from the unconstrained source video and compromising shot-to-shot consistency. We address this with a two-phase approach: Q-Preservation maintains motion structure using Q values from unconstrained generation, while Q-Flow instead preserves feature flow maps, avoiding leaked identity in later steps.

Our main contributions:
1) We provide a systematic analysis of Q-features in text-to-video diffusion models, revealing their dual role in encoding both motion and identity information, with effects persisting longer into the denoising process.
2) We introduce ``\textit{Motion by Queries}'', an efficient approach that enables zero-shot motion transfer in both UNeT and Diffusion-Transformer architectures. %
3) We present a training-free method for consistent multi-shot video generation that balances character consistency and motion quality.

\section{Related work}
\textbf{Motion Transfer in Video Generation:} Motion transfer guides video generation using source video motion patterns, spanning camera movements and object deformations while maintaining structural integrity. Common approaches involve model adaptation (fine-tuning temporal attention~\citep{jeong2024vmc, ren2024customize}, LoRAs~\citep{zhao2024motiondirector}, motion embeddings~\citep{motion_inversion}) or  test-time optimization~\citep{yatim2024space}. \edit{Concurrent noise-warped methods~\cite{lin2025equivdm,burgert2025gowiththeflow} require weeks of training on 8/64 GPUs to inject motion via warped noise, and are out of the scope of our zero-shot setup.}
In contrast, our method enables zero-shot motion transfer via query injection without extra  training or optimization. \edit{In~\cite{ling2025motionclone}, a zero-shot approach is also employed using temporal attention maps, but incurs $\times$12 denoising overhead. Our work analyzes motion in a broader context through spatial Q features, providing principles applicable across video generation tasks and models.}

\begin{figure*}[htbp]
    \centering
    \includegraphics[width=0.75\textwidth,trim={0.1cm 23.2cm 14.7cm 0cm},clip]{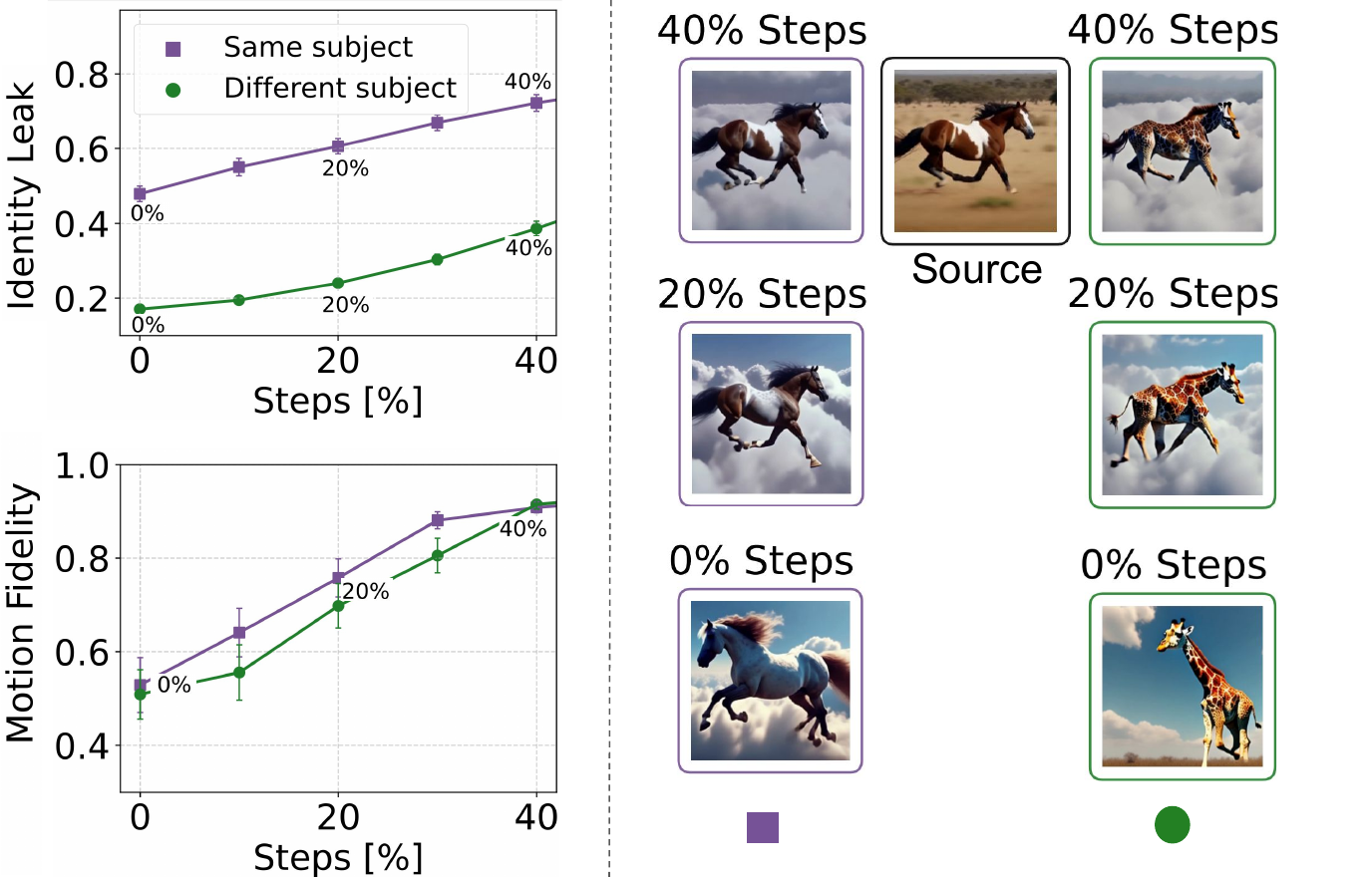} %
    \caption{Same-class motion transfer suffers from identity leakage (purple), worsening with increased Q injection. Cross-class transfer (green) achieves reasonable separation at 40\% injection, where motion quality is also preserved.
The leftmost purple data point shows results for random same-class images (no injection). Quantitative results use motion-transfer data from \cite{motion_inversion}. For Illustration we use frames from the videos in \cref{fig1}. }
    \label{fig_analysis*}
\end{figure*}

\textbf{Video Editing:}
Video editing modifies specific video attributes while preserving others, rather than replicating motion patterns. \citet{bai2024uniedit} also use feature injection for edit and do not address our broader Q injection context.
\citet{zhao2023make} combine ControlNet and CLIP. Image-based methods~\citep{qi2023fatezero, liu2024video, shin2024edit, cong2023flatten, wu2023tune} struggle with coherence, as they lack a temporal model.
\textbf{Structure and Appearance in Text-to-Image Models:}
Early work in text-to-image showed that Keys and Queries guide structure (``Where"), while Values set appearance (``What")~\citet{tewel2023key}.
Further studies established that the layout is set early in denoising~\citep{patashnik2023localizing}, and self-attention queries encode structural layout ~\citep{alaluf2023crossimage, tewel2024training}. We extend these findings to video, showing queries play a more complex role than previously understood - affecting both motion and identity. Additionally, we find that capturing motion patterns in videos requires more denoising steps compared to images, highlighting fundamental differences between these domains.
\textbf{Extended Attention Sharing:}
When using text-to-image models to generate~\citep{wu2023tune,ceylan2023pix2video,Khachatryan_2023_ICCV} or modify a video~\citep{tokenflow2023}, an extended self-attention block~\citep{wu2023tune} is often employed to share keys and values across different frames, enabling them to draw visual appearances from each other and enhance consistency.
Beyond cross-frame consistency, %
it has been used to inject consistent identities from a source image to video~\citep{xu2023magicanimate,hu2023animate,chang2023magicdance,tu2023motioneditor}, maintain appearance in layout editing~\citep{cao_2023_masactrl,avrahami2024diffuhaul}, combine appearances~\citep{alaluf2023crossimage}, for personalization~\citep{gal2024lcmlookahead,zeng2024jedi} and style transfer~\citep{hertz2023StyleAligned}.
\textbf{Consistent generation} aims to maintain consistent subjects across outputs from a generative model, and has mainly studied in text-to-image tasks. A common approach is to use personalization~\citep{gal2022textual,ruiz2022dreambooth} to promote consistency, either through inpainting with a personalized model \citep{jeong2023zero}, personalized LoRA models \citep{simoLoRA2023}, or training LoRAs for large, semi-consistent clusters \citep{avrahami2023chosen}. Alternatively, encoders can inject identity at inference~\citep{ye2023ip,wei_2023_ICCV,gal2023encoder}, but require large pre-training and struggle with new domains. Storyboard-based methods~\citep{feng2023improved,liu2023intelligent}, and video personalization methods \cite{jiang2023videobooth, yuan2024identity, wu2024motionbooth, zhang2025magic} face similar limits.
Recent work ~\citep{tewel2024training,fan2024refdrop}  achieves character consistency without personalization or tuning, using feature sharing across images.

\begin{figure}[htbp]
    \centering

    \includegraphics[width=\columnwidth,trim={0.cm 27.8cm 15.6cm 0cm},clip]{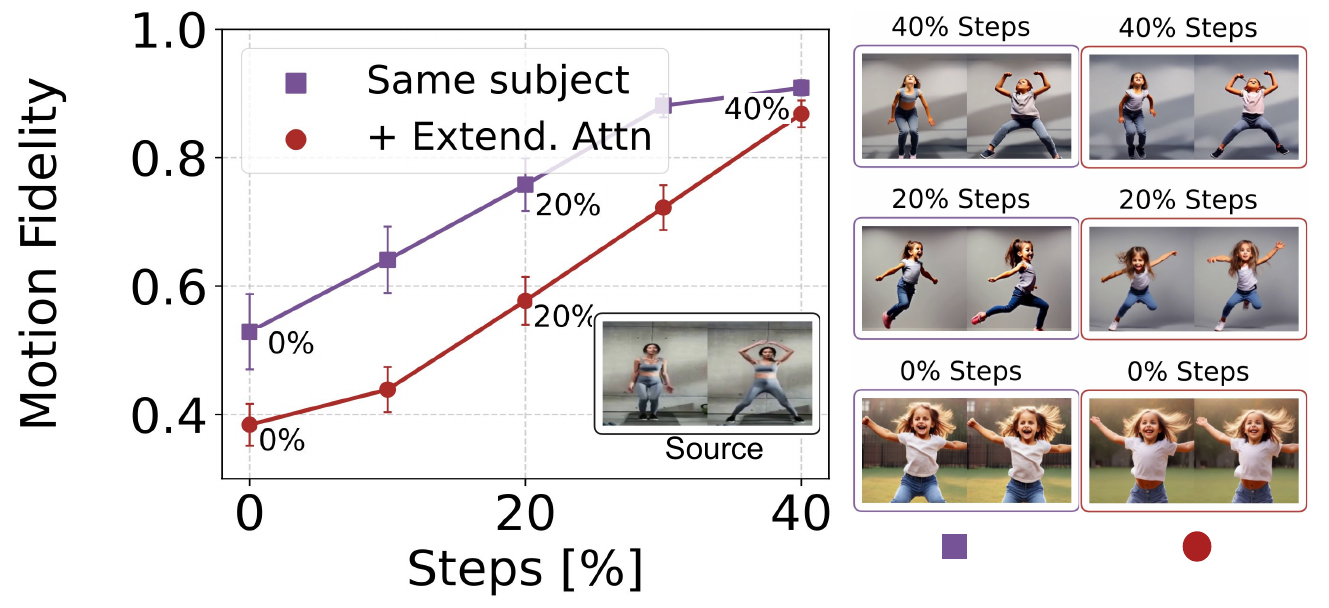}
    \caption{Motion is compromised with extended attention across video-shots. To recover the motion, longer q-injection periods are required, which consequently increases identity leakage. 
    }
    \label{fig_analysis_ex_attn}
\end{figure}

\section{Analysis: Query features in text-to-video generation}
\label{sec_analysis}

To study the role of Q-features in text-to-video (T2V) models, we design an injection experiment. The idea is simple: we take a source video $V_S$ with a known text description  $\tau_S$, and generate a target video $V_T$ using a prompt $\tau_T$. We then record Q features from $V_S$, inject them into the generation of $V_T$, and analyze the effect of that injection. We base our main analysis on VideoCrafter2~\cite{chen2024videocrafter2} model, and later test Q-injection in other architectures for generality.

More specifically, given the video $V_S$, we add noise at a level corresponding to a noisy step $t$, yielding a noisy latent $z_t$. We then perform a single DDPM denoising step (with a $50$-step schedule) and record the  Q features from all \textit{spatial} self-attention layers of the diffusion model. This is repeated $20$ times for various noise levels, resulting in a sequence of $20$ $Q$ tensors ($Q_S(50), Q_S(49), ..., Q_S(30)$, corresponding to DDPM steps $t=1000, 980, ..., 600$).  
Finally, we generate a new video $V_T$ with prompt $\tau_T$, while injecting $Q_S(t)$ tensors at the first $k$ DDPM denoising steps. We vary the amount of steps receiving Q-injection: From none (0\%) up to 40\% of DDPM steps.

To understand the effect of Q-injection, we measure similarity between source and target videos in two aspects: Identity Leakage and Motion Fidelity. 
Identity Leakage measures mean DINO similarity between the frames of source $V_S$ and target $V_T$ videos. Motion Fidelity \cite{yatim2024space} measures cross-correlation between point tracks in source and target videos: 
 \begin{equation*}
     \frac{1}{m}\sum_{\tilde{p}\in \tilde{\mathcal{P}}} \underset{p \in \mathcal{P} }{\text{max}} \ \textbf{corr} (p,\tilde{p})+\frac{1}{n}\sum_{{p}\in {\mathcal{P}}} \underset{\tilde{p} \in \tilde{\mathcal{P}} }{\text{max}} \ \textbf{corr} (p,\tilde{p})
     \label{eq:motion}
\end{equation*}

where $\mathcal{P}=\{p_1,\dots,p_n\},\tilde{\mathcal{P}}=\{\tilde{p}_1,\dots,\tilde{p}_m\}$ are point tracks in source and target videos. \textbf{Error bars:} are standard-error-of-the-mean (S.E.M) to show the significance of our findings.

\cref{fig_analysis*} shows our results. We consider two Q-injection setups: one where the source and target prompts share the same subject (purple), and one where they differ (green). We highlight three key differences in how Q injection behaves in T2V models versus T2I models.

\begin{figure*}[t]
    \centering
    \includegraphics[width=\linewidth, trim={0.cm 13.5cm 1.cm 4cm},clip]{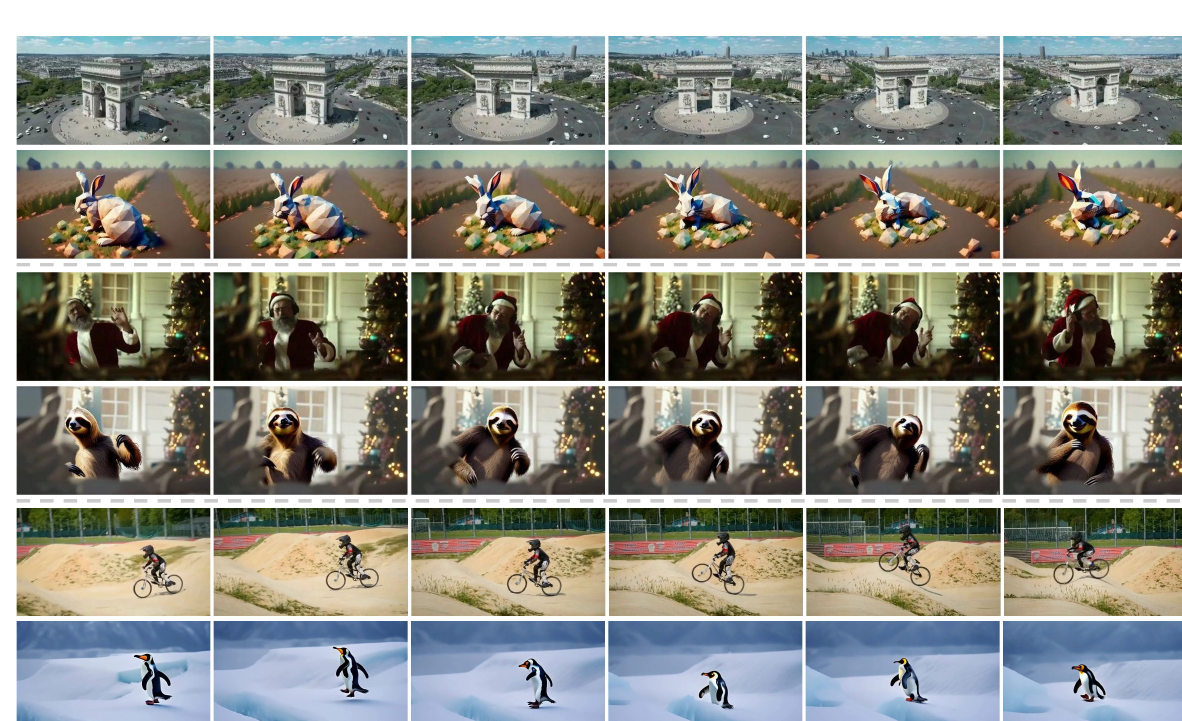} %

    \caption{\textbf{\href{https://research.nvidia.com/labs/par/MotionByQueries/\#fig_qual}{(click-to-view-online)} Qualitative Results, Motion Transfer (VideoCrafter2).} Each pair of rows are frames from source (pair-top) real video and target (pair-bottom) generated video. Transferring Motion by Queries allows to use source videos to inject camera motion (top), non-rigid movement (middle), and combinations of movements (bottom). See the supplemental for videos and more examples. 
    }
    \label{fig_motion_qual}
\end{figure*}

First, \cref{fig_analysis*} (bottom-left), shows that Motion Fidelity increases with Q-injection duration, reaching high similarity at 40\% injection. We find that, unlike images--where structure is established early in the denoising process--videos require significantly more steps to set the motion structure.
Second, more surprising is the top-left panel. It reveals that identity similarity also increases with the duration of injecting Q. This suggests that video generation models also encode identity information into the Q vectors--an intriguing shift from its traditionally assumed role in T2I models.
Third, we observe an interesting phenomenon: when $\tau_S$ and $\tau_T$ use the same subject, the target video often features a subject with an appearance identical to that of the source video while maintaining the background specified in $\tau_T$. This identity ``leakage'' is significantly less pronounced when $\tau_T$ and $\tau_S$ feature different subjects. Qualitatively, in \cref{fig_analysis*} (right), we present a source video of ``A horse galloping in the savanna" with a distinct white spot on its back. Notice that as the number of injection steps increases, the target horse (purple) becomes more similar to the source horse while preserving the cloud background specified in the prompt. On the other hand, when $\tau_T$ describes a giraffe, its shape and motion become more similar to those of the source horse, but its appearance remains a giraffe. This suggests a clear distinction in Q-injection behavior: it transfers appearance and motion when the subjects match, while for different subjects, the transfer primarily affects motion characteristics.

Finally, we examine the effect of Q-injection when the target consists of multiple videos sharing generation features. Feature sharing within a batch is widely used in T2I generation, aiding tasks like consistent character generation~\citep{tewel2024training} and editing~\citep{cao_2023_masactrl}. A common technique is the extended attention mechanism, which modifies self-attention layers to attend across elements of multiple videos in the same batch.
Here, we explore the impact of Q-injection when generating a batch of 3 videos at a time and using shared Extended Attention between shots. In \cref{fig_analysis_ex_attn}, we include the Extended Attention curve to visualize motion fidelity in this scenario.
Our results show that extended attention induces a "motion freeze" effect, requiring Q to be injected for more steps to match the source video's motion. Moreover, since Q-injection in this setting needs more steps, it also introduces identity information, making it difficult to disentangle motion from identity injection.

\noindent\textbf{Generality Across Architectures.}
\edit{We extended our core Q-injection experiment of \cref{fig1} to several T2V models: (1) WAN 2.1~\cite{wan2025}, a SoTA DiT model; (2) T2V-Turbo-V2~\cite{turbo_v2}, a fast sampling model, using as few as 8 denoising steps; and (3) LTX-Video~\cite{HaCohen2024LTXVideo}, a fast DiT model. %
 As in our main findings, Q-injection led to identity leakage for same-subject prompts and primarily transferred motion for different-subject prompts. We also verified that without Q-injection, source and target prompts produced visually distinct outputs. Qualitative results are provided in the supplemental and \cref{fig_teaser_other_models}. See Appendix for implementation details.}

\section{Application 1:  Motion transfer }

Our first application is motion transfer. It allows creators to control motion in a fine way, which cannot be achieved using text. In this problem we are given a video $V_S$ that contain some pattern of movement, and we wish to generate a new video $V_T$ that follows the same movement patterns. The movement can be of entities in the video, or the camera or both.

Our approach, called ``\textit{Motion by Queries}'' follows the experiment described in section 2. First, we extract a series of Q-features from $V_S$, by denoising the model to various timesteps ($t=1000, 980, \ldots ,600$), obtaining $[Q_S(50),\ldots,Q_S(30)]$. 
Then, we inject those Q features during the generation of the target video $V_T$, this time with the new prompt $\tau_T$. 
For WAN 2.1, we extracted Q-features from a single low-noise timestep and injected them into all higher-noise steps during target generation, inspired by~\cite{ling2025motionclone}. See appendix for additional implementation details.

\subsection{Experiments}
\label{sec_experiments_mt}

We evaluate Motion by Queries qualitatively and quantitatively using a Motion-Transfer benchmark \cite{motion_inversion}, comprising 66 video prompts, guided by 22 source videos. Videos were sourced from DAVIS \cite{DAVIS}, WebVID \cite{WebVID}, and online resources, representing diverse scenes, objects, and motion. For the base text-to-video model, we use VideoCrafter2 (VC2).

\textbf{Baselines:}
We compare our approach with recent methods: \textbf{(1) Diffusion Motion Transfer (DMT)}, a test-time optimization approach \cite{yatim2024space}, a \textbf{(2) Motion-Inversion (MI)} \cite{motion_inversion}%
, a fine-tuning method that learns motion-specific temporal embeddings from a source video, and \edit{\textbf{(3) MotionClone (MC)} \cite{ling2025motionclone}, a zero-shot approach where motion is guided by temporal transformer self-attention maps extracted when U-Net processes source videos. }
MI$^*$, MC$^*$ indicate results we reproduced from their online code. Except for one case in the supplemental, all qualitative comparisons use videos from MI and DMT project pages.
We also adopt from \cite{motion_inversion} the quantitative results of Video Motion Customization (VMC) \cite{jeong2024vmc} and Motion-Director (MD) \cite{zhao2024motiondirector}, both are fine-tuning-based approaches.

\textbf{Evaluation Metrics:}
\edit{We measure Motion Fidelity (M. Fidel.), the point track correlation between generated and source videos, as described in Sec.~\ref{sec_analysis}; Temporal Flicker (T. Flick.) which quantifies flickering artifacts, using mean absolute error between consecutive frames, as in VBench~\cite{vbench}; Text Similarity (Text), the average CLIP-Text score between generated frames and the textual prompt~\cite{radford2021learning}; Temporal Consistency (Temp. C.) which measures video coherence by averaging CLIP-Image similarity between consecutive frames ~\cite{vbench}; Identity Leak (Id. Leak) to quantify the similarity between source and target video frames, as described in Sec.~\ref{sec_analysis}; Finally, Aesth., Smooth., and Bk Cons. which are VBench quality metrics for visual appeal, temporal smoothness, and background consistency.}

\begin{figure*}[t]
    \centering
    \includegraphics[width=0.87\linewidth, trim={0.cm 22.4cm 12.cm 0.cm},clip]{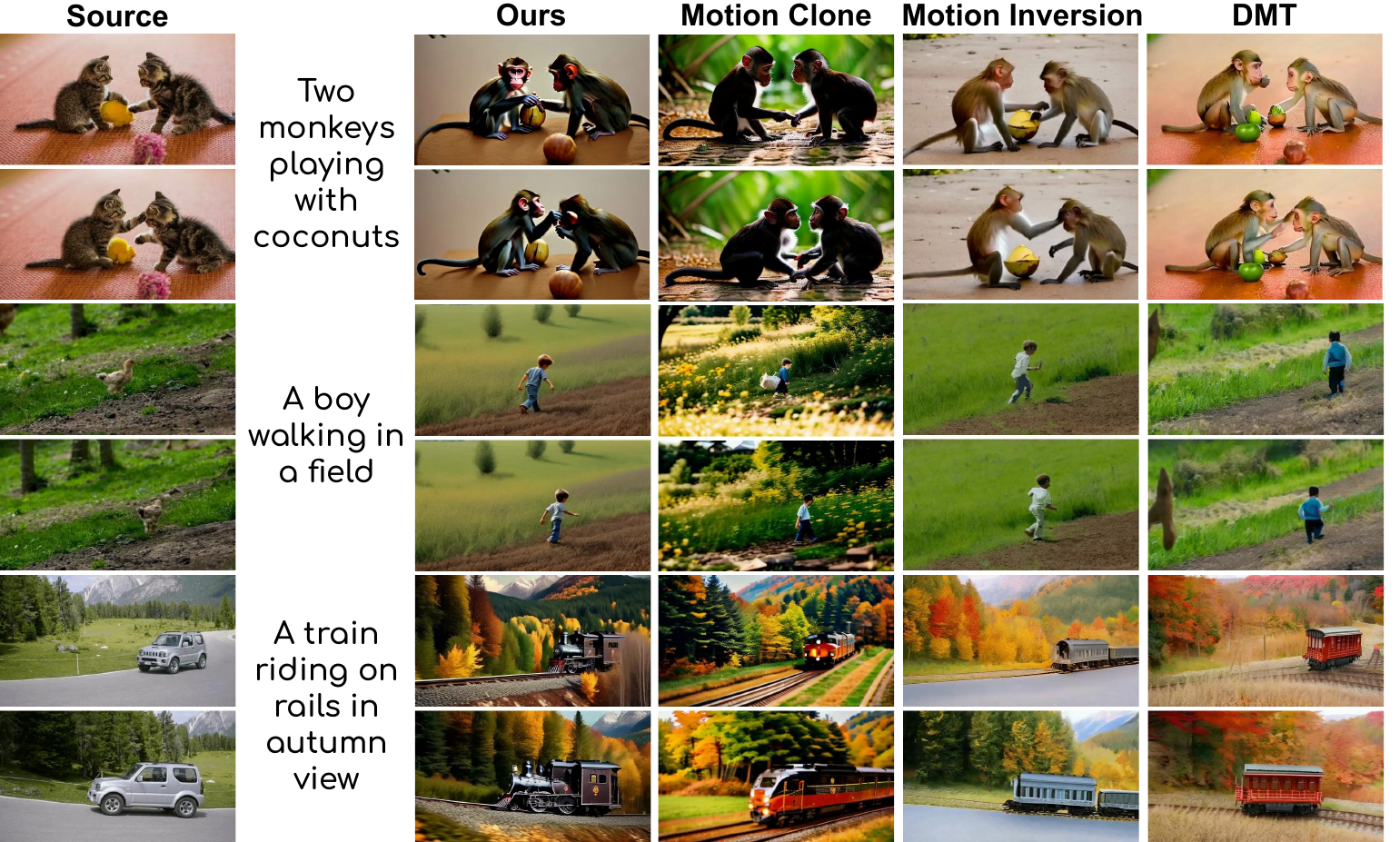} %

    \caption{\textbf{\href{https://research.nvidia.com/labs/par/MotionByQueries/\#fig_compare}{(click-to-view-online)} Qualitative Comparisons, Motion Transfer.} 
        }
    \label{fig:motion_compare}
\end{figure*}

Additionally, we compare runtimes and overhead relative to each method's base model.  We break the runtime to components: ``Invers.'' is for  inversion or feature recording from source video,
``Optim.'' for optimization or tuning,
``Infer'' for sampling a new video,
``Sum'' is the total runtime, and ``Overhead'' is the ratio of ``Sum'' to inference in the base model. Runtimes were measured on NVIDIA H100, when generating at a resolution of 576$\times$320.

\textbf{Quantitative Results:}
Our approach demonstrates competitive results across various metrics while remaining simple and optimization-free (Tables \ref{table_motion_metrics}, \ref{table_motion_runtime}).
Specifically, our method achieves lower (better) identity leakage (38.6) than MI$^*$ (43.7), with text alignment (28.8) and temporal consistency (97.0) comparable to baselines. For motion fidelity (MF), our method reaches 91.5, which, while lower than MI$^*$ (97.0), fidelity remains sufficient for some practical scenarios as supported by qualitative results. \edit{Compared to MC, our method attains higher identity leakage (38.6 vs. 24.2, lower is better), while achieving improved temporal stability, as reflected in a higher flickering score (95.1 vs. 86.0).}
In terms of efficiency, our method requires only $\times$1.2 overhead (70s) relative to the base model (VC2), \edit{and is more efficient than MC ($\times$12, 104s), MI ($\times$23, 208s), and DMT ($\times$45, 410s).}

\begin{table*}[htbp]{
    \begin{sc}
    \scalebox{.94}{
    \renewcommand{\arraystretch}{.8}
       
    \begin{tabular}{lcccccccc}
    \toprule
 & M. Fidel. $\uparrow$ & T. Flick. $\uparrow$ & Text $\uparrow$ & Temp. C.  $\uparrow$ & Id. Leak $\downarrow$ & Aesth. $\uparrow$ & Smooth. $\uparrow$ & Bk Cons. $\uparrow$ \\

    \midrule
    DMT& 78.8& - & 28.8& 93.6 & - & -& -& - \\ 
    VMC& 93.7& - & 27.1& 94.6 & - & -& -& - \\ 
    MD& 93.9& - & 30.4& 93.3 & - & -& -& - \\ 
    MI& 95.5 & - &  31.1& 93.5 & - & -& -& - \\ 
    \midrule

MI$^*$ & \textbf{97.0 $\pm$ 0.4} & 92.2 $\pm$ 0.4 & 29.2 $\pm$ 0.6 & 96.8 $\pm$ 0.2 & 43.7 $\pm$ 1.5 & 52.6 $\pm$ 1.2 & 95.6 $\pm$ 0.3 & 94.5 $\pm$ 0.3 \\

MC$^*$ & 95.0 $\pm$ 0.7 & 86.0 $\pm$ 0.6 & \textbf{29.9 $\pm$ 0.4} & 95.8 $\pm$ 0.3 & \textbf{24.2 $\pm$ 1.1} & \textbf{55.6 $\pm$ 0.9} & 93.5 $\pm$ 0.4 & 93.3 $\pm$ 0.4 \\

Ours & 91.6 $\pm$ 0.9 & \textbf{95.1 $\pm$ 0.3} & 28.8 $\pm$ 0.6 & \textbf{97.0 $\pm$ 0.2} & 38.6 $\pm$ 1.8 & 54.1 $\pm$ 1.3 & \textbf{97.3 $\pm$ 0.2} & \textbf{94.9 $\pm$ 0.4} \\

    \bottomrule
    \end{tabular}
    
    }
    \end{sc}
    
    \vspace{2pt}
    \caption{\textbf{Quantitative Evaluation Metrics, Motion Transfer.} Values are mean $\pm$  S.E.M.}
    \label{table_motion_metrics}
    }
\end{table*}

\begin{table*}[h]
    \centering
    \begin{minipage}{0.7\textwidth}        
\centering
    \begin{sc}
    \scalebox{0.94}{
    \renewcommand{\arraystretch}{.8}
       
\begin{tabular}{lccccc}
\toprule
 Time \small[sec.\small] & Invers. & Optim. & Infer & Sum & \textbf{Overhead} $\downarrow$ \\

\midrule
\textbf{Z.scope} & - & -  & 9 & 9 & - \\
DMT & 260  & & 150  & 410 & $\times$45 \\

MI & 9 & 190  & 9 & 208 & $\times$23 \\
\midrule
\textbf{AnimateDiff} & - & -  & 9 & 9  & - \\
MC  & 0.3 & 0  & 104 & 104 & $\times$12 \\
\midrule

\textbf{VC2} & - & -  & 58 & 58  & - \\
Ours  & 12 & 0  & 58 & 70 & \textbf{$\times$1.2} \\
\bottomrule
\end{tabular}

    }
    \end{sc}

    \end{minipage}%
    \begin{minipage}{0.29\textwidth}         
    \caption{\textbf{Runtime Comparison.} 
    Existing methods are $\times$12-45 slower than their base model. Ours adds merely $\times$1.2 overhead.
    }
    \label{table_motion_runtime}    
    \end{minipage}

\end{table*}

\begin{figure*}[t]
    \centering
    \includegraphics[width=\linewidth, trim={0.cm 16.25cm 1.cm 0cm},clip]{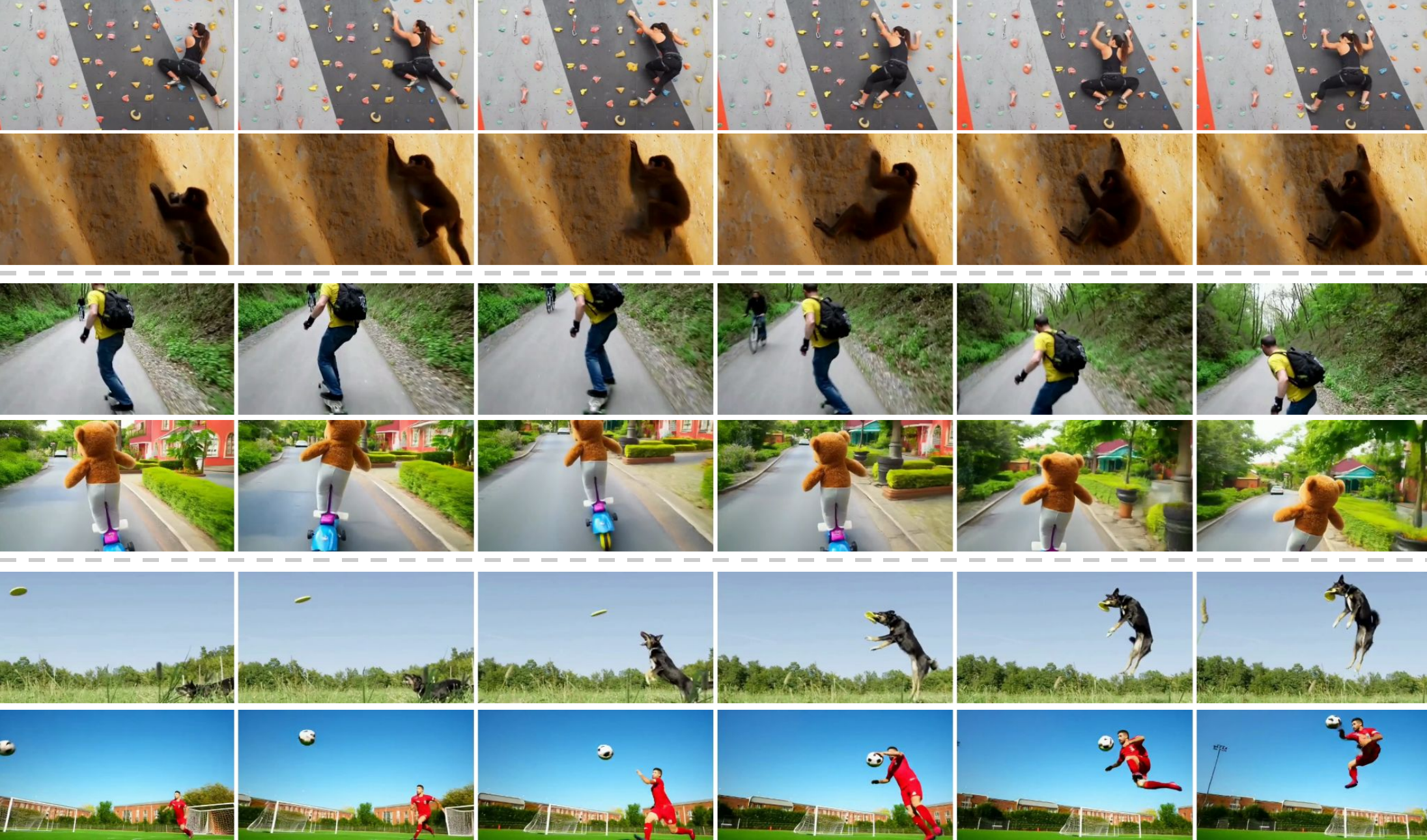} %

    \caption{\textbf{\href{https://research.nvidia.com/labs/par/MotionByQueries/\#fig_more_qual_wan}{(click-to-view-online)}}\textbf{ Qualitative Results with Wan 2.1, Motion Transfer.} Each row pair shows source (top) and generated (bottom) frames. %
    See supplemental for videos and more examples.
    }
    \label{fig_qual_wan_main}
\end{figure*}

\textbf{Qualitative Results:}
In Figure \ref{fig_motion_qual} and supplemental, we show that Motion by Queries guides videos generated by VC2 to follow the motion in source videos, capturing both camera motion (top), and non-rigid object movement (bottom). Figure \ref{fig:motion_compare} compares our results with MC, MI, and DMT: in the monkeys example, our method produces motion that resembles the source video, albeit with slightly lower motion magnitude than the baselines. For the boy and train examples, our results are comparable to those of MI and DMT. Supplemental comparisons further reveal cases where \edit{MC exhibits temporal instability} and MI identity leakage, as seen in the quantitative results. They also provide examples illustrating custom camera motion and hybrid object-camera motion.
\edit{Figure \ref{fig_qual_wan_main} and supplemental present results with the WAN 2.1 DiT model. Compared to VC2, WAN exhibits a stronger physical prior: it tends to generate motion that is physically plausible and realistic, sometimes prioritizing physical accuracy over exact pixel-level replication. This is noticeable in non-rigid motion scenarios, such as the wall climbing example. We also find that WAN is more sensitive to the choice of initial noise, resulting in greater variability in the generated outputs.}

\section{Application 2:  Consistent multi-shot video generation}
\label{sec:consistent_video_main}

Generating long, coherent sequences remains difficult for text-to-video models, yet this ability is key for rich storytelling and creative video applications. A practical approach is to generate several short video shots with the same characters, as in cinematic videos. The main challenge is keeping character consistency across shots, since current models excel at single clips but struggle with cross-shot generation.

Prior work~\citep{tewel2024training,fan2024refdrop} tackled character consistency in text-to-image by sharing self-attention features. However, as our analysis shows, video features encode both identity and motion, making it challenging to apply these methods to video, where naive extended attention can cause synchronized or diminished motion.
To address this, we propose a method tailored for consistent video generation, combining extended attention for consistency~\cite{tewel2024training} with a two-step Q injection process. Early Q preservation sets motion structure, followed by flow-based Q injection that lets Q values evolve for better consistency while maintaining structure.

\begin{figure*}[h]
    \centering
    \makebox[\textwidth][c]{%
        \hspace{-.1cm}%
        \begin{minipage}[t]{8.cm}
            \includegraphics[width=\linewidth, trim={0.1cm 0cm 0cm 0cm},clip]{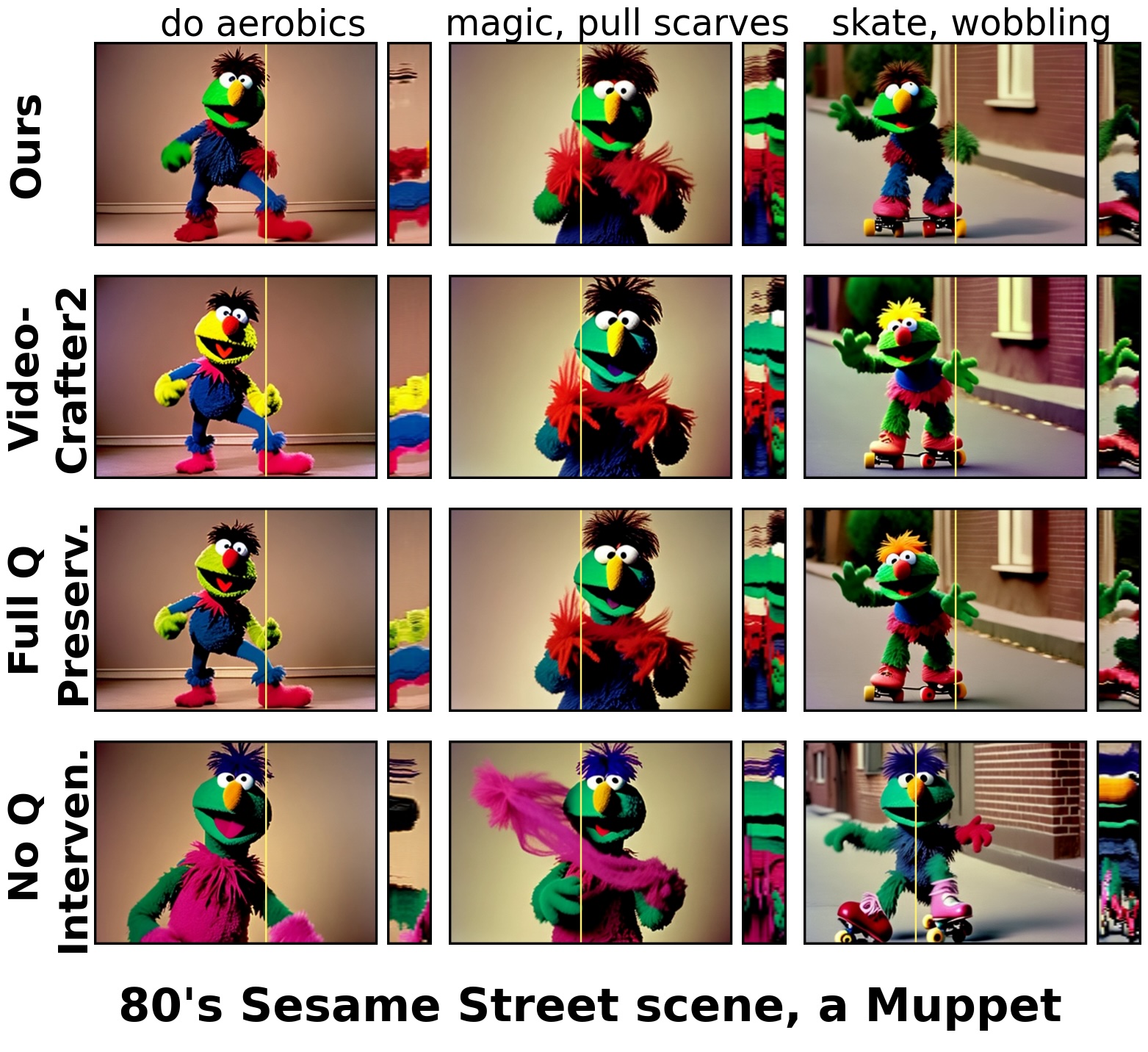}
        \end{minipage}%
        \hspace{0.3cm}%
        \begin{minipage}[t]{7.45cm}
            \includegraphics[width=\linewidth, trim={3.07cm 0.1cm 0cm 0cm},clip]{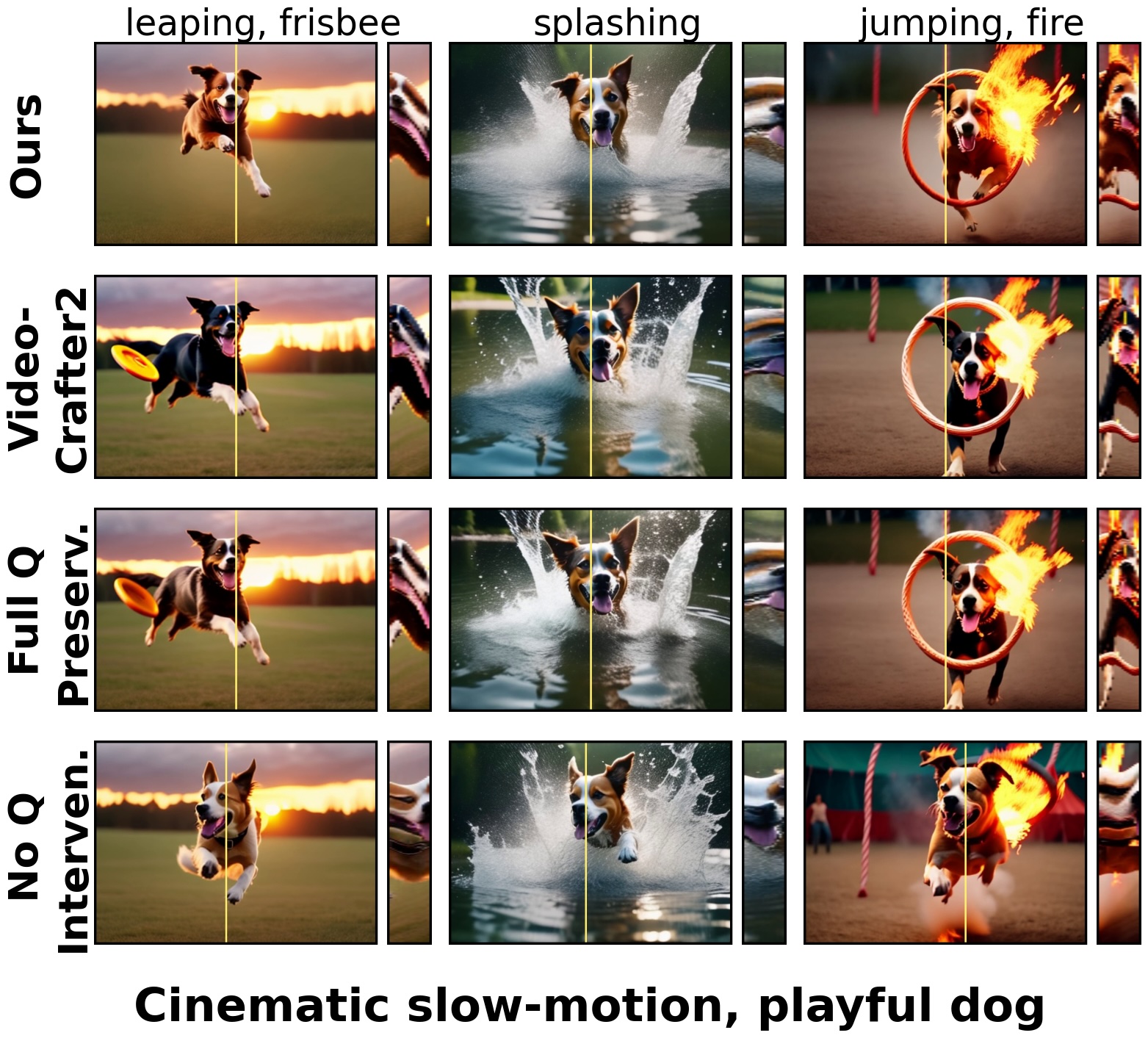}
        \end{minipage}%
    }
\caption{\textbf{\href{https://research.nvidia.com/labs/par/MotionByQueries/\#ablation_Q}{(click-to-view-online)} Comparing Q token intervention strategies for consistent video generation.} ``Ours" (top row) balances character consistency and natural motion. VideoCrafter2 (second row) offers diverse motion but doesn't allow for character consistency. ``Full Q Preservation" (third row) without flow-based processing, preserving original motion but losing character consistency since identity leaks from VideoCrafter2. ``No Q Intervention" (bottom row) maintains character consistency but suffers from motion degradation and synchronization across shots. The right side of each example shows y-t slices (temporal cross-sections) along the yellow vertical line visible in each frame, revealing motion patterns over time.}
    \label{fig_ablation_Q}
\end{figure*}

We build on ConsiStory~\cite{tewel2024training} (Appendix ~\ref{sec:consistory_details}) that applies extended attention within a subject mask to share K, V self-attention features, but reduces layout variability. To restore diversity, ConsiStory injects Q features from a vanilla, non-consistent sampling step. While this works for images, injecting Q features from vanilla videos preserves motion diversity but causes identity leakage from those videos, breaking character consistency (\cref{fig_analysis_ex_attn}).

Our two-stage solution—Q preservation followed by Q Flow—addresses this. In early diffusion steps, which primarily control motion and layout, we inject the Q-features from vanilla videos generated without extended attention.
Then, in the subsequent \textit{Q-Flow} phase
we apply a relaxed Q-injection that preserves the flow of Q-features rather than injecting the features directly. This technique is inspired by TokenFlow~\cite{tokenflow2023}. %
 Specifically, for each vanilla video, we first compute the nearest-neighbor correspondence field on its Q-features, defining a Q-feature flow to maintain in the generated consistent video. We then inject this correspondence into our generated video (see Appendix~\ref{formal_flow_Q} for details).

\subsection{Experiments}
\label{sec::consis_exp_main}
We investigated the effects of self-attention query (Q) tokens on motion and identity for consistent video generation described above. Further extensive evaluations, including baselines, user studies, and ablations, are in  Appendix~\ref{sec::consistent_video_sup}.
\cref{fig_ablation_Q} shows typical generations for different Q token strategies when combined with extended self-attention.%

Without Q intervention (\cref{fig_ablation_Q}, 4th row), subject identity is preserved across shots, but motion quality drops: %
(1) movements become synced, \eg, the dog's head turning simultaneously in all shots. (2) motion style and pose vary less: repeatition across shots, \eg, the dog's leap, the Muppet's centered swaying, the camera movement becomes static in the skating Muppet shot. (3) motion-artifacts, to reconcile motion sharing with the varying text prompts, \eg, the skating Muppet's body appears frozen while its legs are displaced.
In contrast, Injecting Q tokens from vanilla videos (3rd row) restores motion but loses subject identity, \eg the Muppet's colors revert to those of the vanilla model.
This reiterate the dual role of Q tokens: injecting vanilla Qs \textit{restores motion}, but leaks the vanilla non-consistent identities into the multi-shot mechanism. %
Our approach (1st row) balances both, restoring the original motion, including nuanced details like body and face orientations, postures, specific body parts %
, as well as parallax camera movement.

\vspace{-5pt}
\section{Conclusion and Limitations}

Our findings show that, unlike in text-to-image models, Q-injection in video models affects both motion and identity, requiring more denoising steps to establish motion patterns. This challenges the classic separation of "where" and "what" pathways~\cite{tewel2023key,patashnik2023localizing,alaluf2023crossimage}, revealing a more complex interplay in video models.
These insights led to two applications: zero-shot motion transfer and training-free consistent multi-shot video generation. 
Our method has limits: motion fidelity can drop with fast movements, and cross-class transfer may cause shape leakage (See \cref{fig_limitations} in Appendix). %
In multi-shot generation, balancing identity and motion remains hard—Q injection can still hurt identity, and partial dropout may be needed to preserve motion (\secref{suppl_q_dropout}).
\clearpage
\newpage

\newpage
\bibliography{main}
\bibliographystyle{icml2025}

\newpage
\appendix
\onecolumn
\section{Appendix}

\begin{figure}[htbp]
     \centering
    \includegraphics[width=13cm,trim={0.cm 3cm 4.5cm 0.cm},clip]{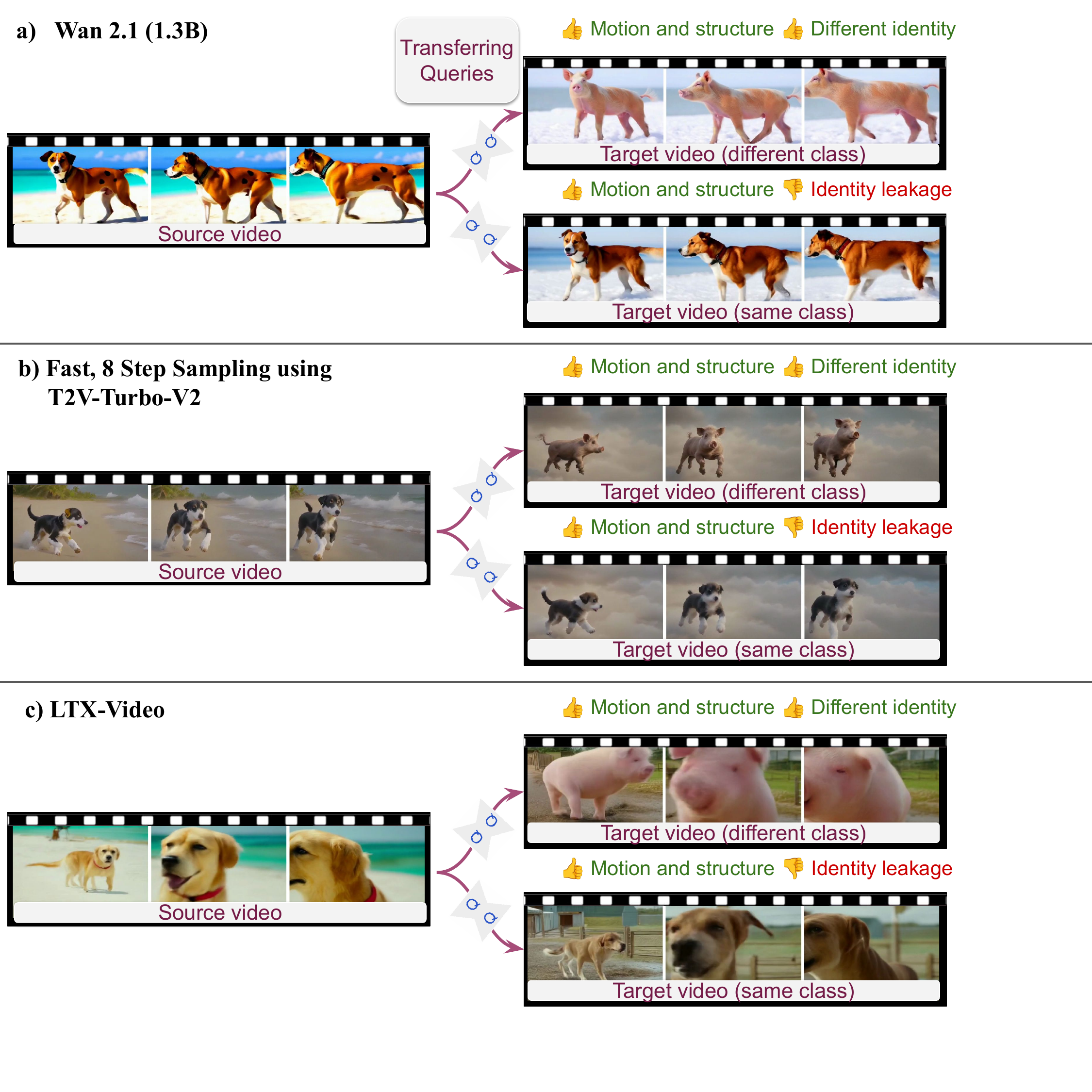} %
    \caption{ \textbf{\href{https://research.nvidia.com/labs/par/MotionByQueries/\#teaser}{(click-to-view-online)}} \textbf{Identity Leakage and Motion Transfer with Other Text-to-Video Models.}
    } 
    \label{fig_teaser_other_models}
\end{figure}

\begin{figure}[htbp]
     \centering
    \includegraphics[width=6cm,trim={0.cm 15.7cm 0cm 0.17cm},clip]{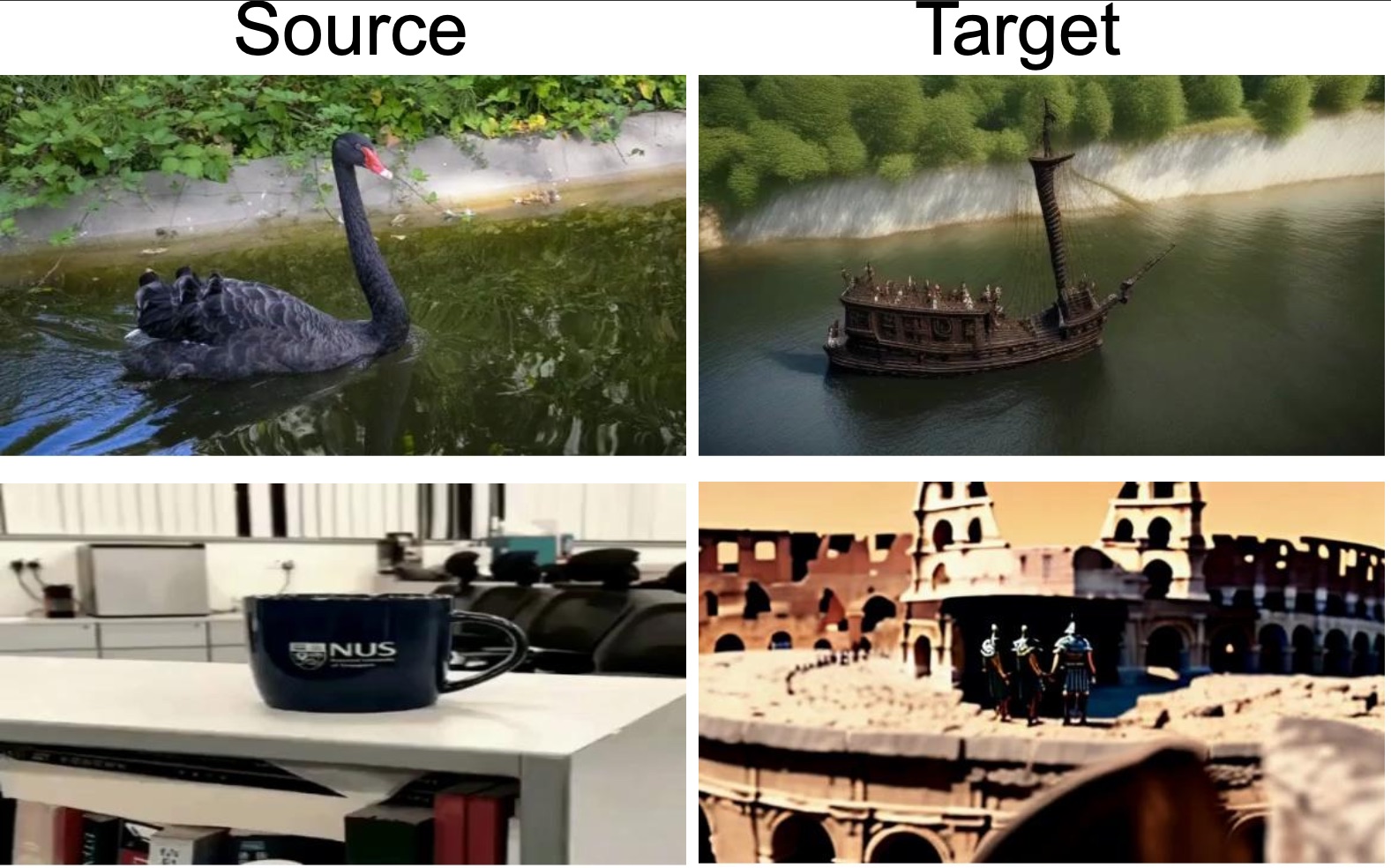} %
    \caption{ \textbf{Limitations.} The source subject shape may affects the target object.
    } 
    \label{fig_limitations}
\end{figure}

\subsection{Motion Transfer Implementation Details}
\label{suppl_mt_details}

\begin{itemize}
    \item We record and inject only the Q features from the conditional forward pass.
    \item When recording Q features, we use the same noise seed across different noise levels. During generation, however, we start from a random initialization.
    \item In VideoCrafter2, to ensure consistency with the video model's latent space, we follow \cite{meiri2023fixed}: we add minor noise to the source video latent (two steps) and then denoise it back to a clean latent before extracting Q features.
    \item For quantitative results, we used Q injection ending at $t=600$. For qualitative results with VideoCrafter2, we selected videos with Q injection ending between $t=800$ and $t=600$: closer to $t=800$ for camera motion, and closer to $t=600$ for non-rigid movement.
\end{itemize}

\subsubsection{DiT-Based Models}
\begin{itemize}
    \item For both WAN and LTX-Video, we used a 50-step flow-matching denoising scheduler as described by Esser et al.~\cite{sd3}. This schedule shifts the timestep allocation so that more steps are concentrated in the high-noise region. Specifically, we used the FlowMatchEulerDiscreteScheduler from huggingface, with their default $\mu=3.065$ hyperparam.
    \item Similar to VideoCrafter2, to transfer the full magnitude of motion we had to inject Q features for a substantial amount of steps. For WAN, we injected Q features for $58\%$ or $60\%$ of the denoising schedule; for LTX-Video, we used $40\%$.
    \item In WAN, we inject Q features only in layers 20--30. In all other models, Q injection is applied to all layers.
    \item Current motion transfer benchmarks consist of short 16- or 24-frame videos, which is significantly shorter than the standard length of WAN 2.1 videos (81 frames).  Therefore, for motion transfer in WAN, we repeat each frame twice and pad the last frame one additional time, therefore mapping $16 \rightarrow 33$ and $24 \rightarrow 49$. Accordingly, we double the frame rate of the generated videos to keep the duration the same as in the source video.
    \item For LTX-Video only, we found that injecting Q features between different initial noise seeds preserved the findings about identity leakage but introduced visual artifacts. To mitigate this, we used identical seeds for different prompts and matched the global statistical moments of the Value features, improving compatibility with the injected Query features.
\end{itemize}

\subsection{Background: Self-Attention in \TTV models}
\label{sec_suppl_background}
Our method manipulates the activations of the spatial self-attention in \TTV diffusion models. We start by outlining its mechanism and introducing key notations.

Recent \TTV diffusion models are based on a latent video diffusion model (LVDM) architecture where a U-Net denoiser is trained to estimate the noise in the noisy latent codes input. The denoising U-Net is a 3D U-Net architecture consisting of a stack spatio-temporal blocks comprised of convolutional layers, spatial transformers (ST), and temporal transformers (TT). The ST operate independently on each video frame, without awareness of the temporal structure, while the TT operate independently on each temporal patch, without awareness of the spatial structure. In this work, we focus on manipulating the self-attention mechanism of the spatial transformer layers.

\section{Consistent Video Generation - Supplementary Detalis}\label{sec::consistent_video_sup}
\subsection{Notations}
\label{sec_notation}
Our method manipulates spatial self-attention activations in \TTV diffusion models. We denote by $\{Q, K, V, O\}$ the respective Query, Key, Value and Output features of a single self-attention layer (see Appendix \ref{sec_suppl_background} for background). In our method, these features interact across frames, enabling cross-frame attention and consistency. We denote by $Q_v$ the $Q$ features of a layer during a ``vanilla'', non-consistent, forward pass in a pretrained network, $Q_c$ the query features from our subject-consistent model, and $Q_f$ as the flow-based query features.
For brevity, we omitted the frame index $i$

\subsection{ConsiStory details} \label{sec:consistory_details}
ConsiStory \cite{tewel2024training} operates in three steps. \textbf{(1) Subject-Driven localization with extended Self-Attention (SDSA) -- } localizes the subject across a set of noisy generated images by aggregating cross-attention maps across layers and timesteps. To ensure subject consistency, SDSA enables each image to attend to patches of the main subject present in \textit{other} image frames. This is done by extending the self-attention mechanism, allowing it to share K, V features of the subject between multiple images.  Unfortunately, SDSA alone diminishes \textit{layout} diversity in the generated images. Therefore, \textbf{(2) Layout Diversity -- } reinforces diversity through two techniques: First, it incorporates Q features from a vanilla, \textit{non-consistent} sampling step. Second, it applies an inference-time dropout to the shared K, V features.
Finally, \textbf{(3) Refinement Injection -- } improves consistency in finer details by injecting the O features between corresponding subject patches.

\begin{figure}[t]
     \centering
    \includegraphics[width=\linewidth, trim={0.cm 10cm 0cm 0cm},clip]{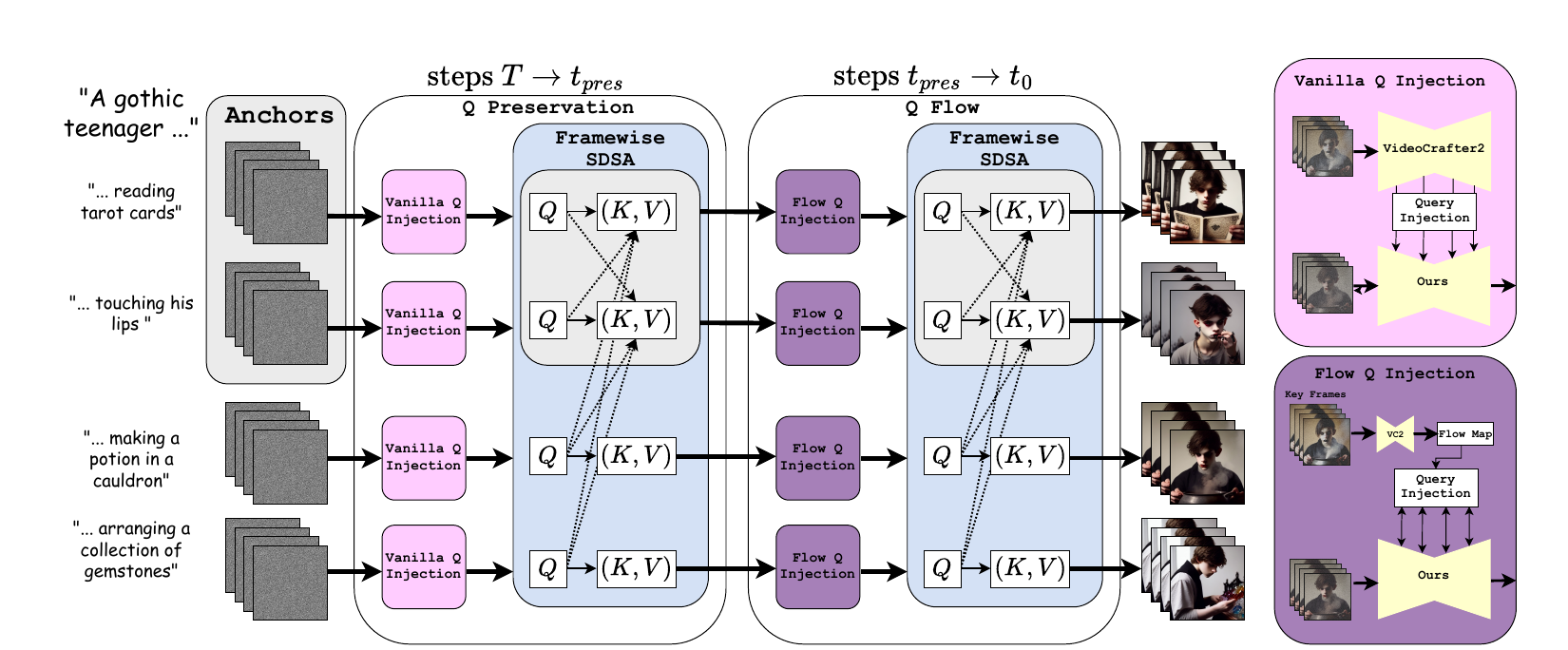} %
    \caption{ \textbf{\ourmethod{} Architecture:} Our consistent denoising process has two phases: Q Preservation and Q Flow. We first generate and cache video shots using ``vanilla'' VideoCrafter2. In Q Preservation ($T \rightarrow t_{pres}$), we use Vanilla Q Injection to maintain motion structure by replacing our Q values with vanilla ones. In Q Flow ($t_{pres} \rightarrow t_{0}$), we use a flow map from vanilla key frames to guide Q feature injection. This phase maintains character identity by allowing the use of Q features from our consistent denoising process, while the flow map ensures that these identity-preserving features are applied in a way that's consistent with the original motion. Throughout, we employ two complementary techniques: framewise subject-driven self-attention for visual coherence, and refinement feature injection (\secref{sec_refinment_FI}) to reinforce character consistency across diverse prompts.
    } 
    \label{fig_architecture}
\end{figure}

The pipeline is illustrated in \cref{fig_architecture}.

\subsection{Framewise Subject-Driven Self-Attention  }\label{sec:extended_attention} %

Our first step builds on the Subject-Driven Self-Attention (SDSA) mechanism \cite{tewel2024training} %
to incorporate subject features across multiple video shots by extending the self-attention mechanism. We identified two critical challenges when adapting SDSA to video generation: (1) reliably localizing the subject during video denoising, and (2) ensuring motion fluidity is not compromised.

For subject localization, we propose using the estimated clean image $\hat{x_0}$ for mask generation instead of relying on internal network activations, ensuring reliable masks even in early denoising steps. For motion fluidity, we introduce a framewise attention scheme, where frames with matching temporal indices across shots selectively attend each other. This prevents artifacts and frozen motion.

We term this component Framewise-SDSA. Further technical details, including the mask estimation process and the formal definition of Framewise-SDSA, are provided in Appendix~\ref{fSDSA_formal}..

When generating multiple video shots with consistent subjects, we face a fundamental trade-off between subject consistency and motion quality. Our experiments show that while Framewise-SDSA improves subject consistency, it often results in side-effects, leading to excessive synchronization of motion layout across video shots and introduces motion artifacts (\cref{fig_ablation_Q}(4th row)). These artifacts arise from the model's attempt to simultaneously satisfy both the text prompt and the undesired synchronization across shots.

Prior work in ConsiStory (Sec. ~\ref{sec:consistory_details}) demonstrated success in maintaining layout diversity for image generation through SDSA dropout and query injection. However, our experiments show that directly extending this approach to video generation produces poor results, with significant visual artifacts and compromised consistency between shots (\cref{fig_ablation_cs}). This likely occurs because (1) Consistory's query injection is applied for shorter periods compared to the amount required in video models, and (2) since ConsiStory's vanilla-network queries are derived from latents that are influenced by consistency-preserving mechanisms in earlier steps, rather than following an independent denoising trajectory.

Our analysis (\cref{fig_ablation_Q}) reveals that query features encode both motion patterns and subject identity. Injecting only vanilla query features ($Q_v$) preserves dynamic motion but results in inconsistent subjects across shots (row 3). Conversely, using only consistency-aware query features ($Q_c$) ensures subject consistency but produces rigid, unnatural, and synchronized movements (row 4). This observation motivates our two-phase approach that leverages both feature types.

\textbf{Phase 1: Motion Structure Establishment.} In early denoising steps ($t\in[T, t_{\text{pres}}]$), we focus on establishing a robust initial motion structure using a process we call Q Preservation. During this phase, we  directly inject vanilla query features ($Q_v$) from pre-generated video shots. This allows us to retain the motion patterns present in the vanilla videos.  Without this initial phase, later denoising steps may deviate from the original motion patterns, leading to degraded motion quality.

\textbf{Phase 2: Flow-based Consistency Integration.}
As denoising progresses (beyond $t_{\text{pres}}$), subject consistency becomes increasingly important. To address this, we introduce Q Flow, a technique inspired by TokenFlow \cite{tokenflow2023}, where flow-based query features ($Q_f$) are injected to incorporate subject-consistent information while preserving the original motion. Similar to \cite{tokenflow2023}, in this phase, we derive a flow map from vanilla-generated keyframes ($Q_v$), which provides the motion structure. We then blend subject-consistent query features ($Q_c$) from nearby frames, as dictated by the flow. This blending process produces $Q_f$, that adhere to the original motion patterns while maintaining subject consistency across frames.

By following this approach, we maintain the natural flow of motion established in Phase 1 and progressively integrate subject-consistent features without sacrificing motion quality. The formal definition of our flow-based query injection process is provided in Appendix \ref{formal_flow_Q}.

\subsection{Refinement Feature Injection for Enhanced Consistency} %
\label{sec_refinment_FI}
Despite improved motion preservation and subject consistency, fine details in subject appearance can still vary across frames. We address this by adapting the refinement feature injection technique.

However, naively applying refinement feature injection solely to the conditional denoising step, as in ConsiStory, introduces unnatural motion artifacts. This is likely due to the conditional step uses a correspondence map to inject features from different frames, while the unconditional step does not, resulting in inconsistent feature injection.
To mitigate this, we extend refinement feature injection to the unconditional denoising step, using the same DIFT correspondence map. We also utilize the entire frame set of each anchor video for refinement injection. This synchronized approach improves overall consistency and reduces motion artifacts. For qualitative results, see \cref{fig_ablation_cs}.

\subsection{Consistent Video Generation - Comparisons to Baselines}

\begin{figure}[t]
    \centering
    \makebox[\textwidth][c]{%
        \hspace{-.1cm}%
        \begin{minipage}[t]{6.5cm}
            \includegraphics[width=\linewidth, trim={0.1cm 0cm 0cm 0cm},clip]{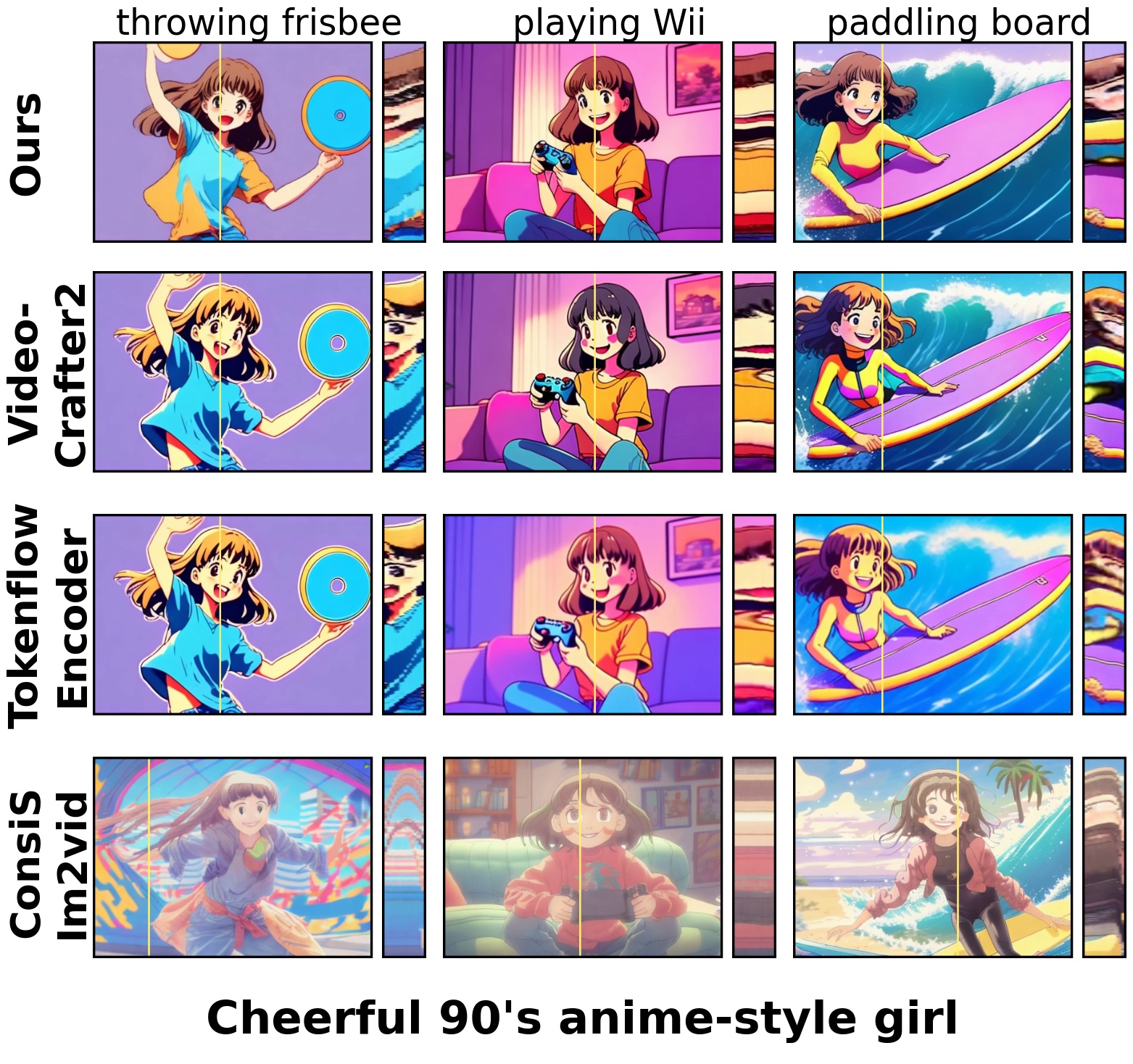}
        \end{minipage}%
        \hspace{0.3cm}%
        \begin{minipage}[t]{7.45cm}
            \includegraphics[width=\linewidth, trim={3.07cm 0cm 0cm 0cm},clip]{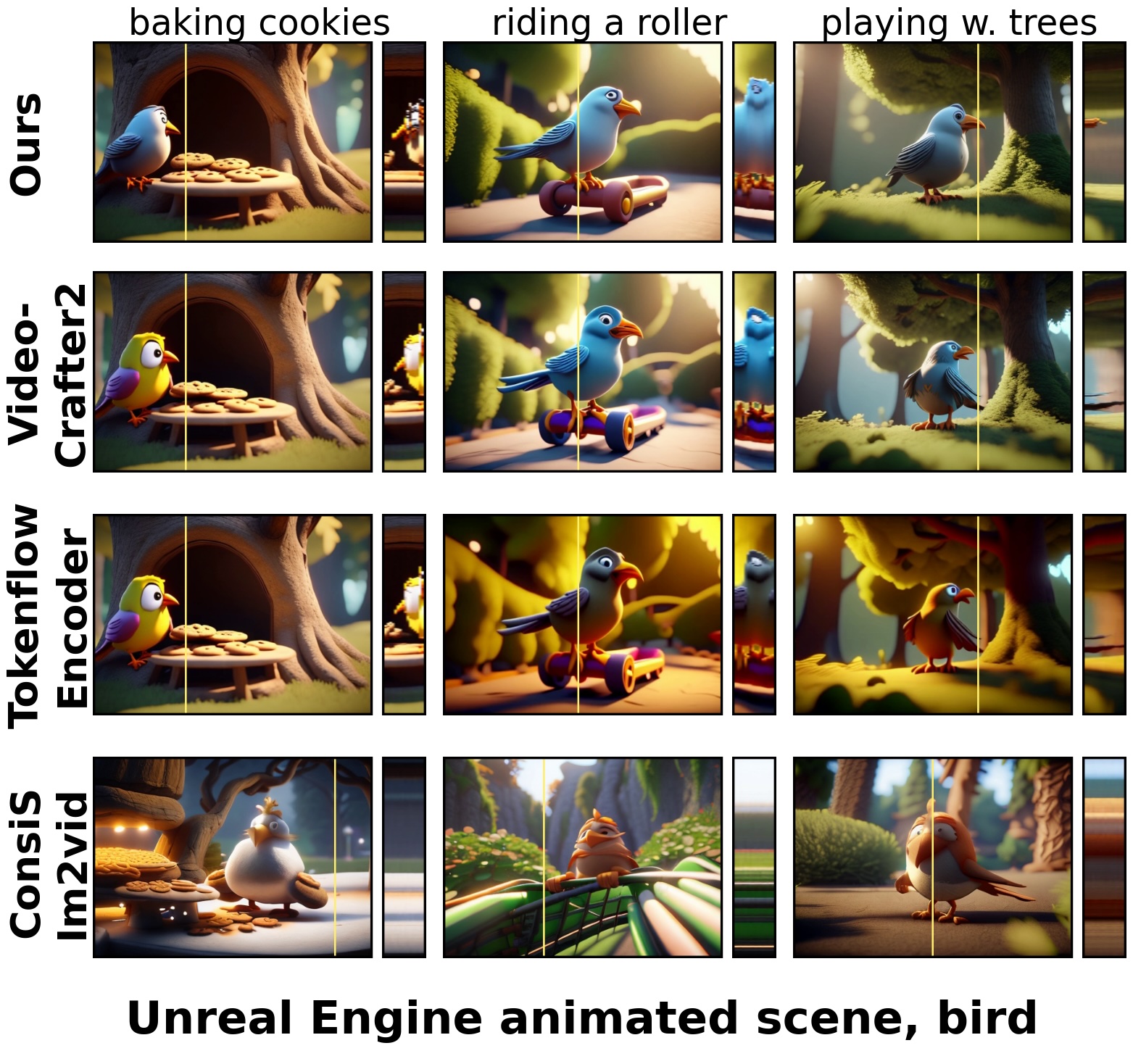}
        \end{minipage}%
    }
    \caption{\textbf{Qualitative Comparisons.} The first frame of each video shot is displayed along with a spatiotemporal y--t slice to visualize motion. 
    \textit{Ours (top row)} shows improved character consistency across shots while maintaining natural motion. \textit{VideoCrafter2} (row 2) is the vanilla model, showing diverse motion but inconsistent characters. \textit{Tokenflow-Encoder} (row 3) preserves original motion but struggles with character consistency and introduces coloring artifacts. \textit{ConsiS Im2Vid (bottom row)}  fails to maintain consistency and exhibits limited motion adherence to prompts. See more examples in \cref{fig_qual_extra}.
    }
    \label{fig_qual}
\end{figure}
We compare \ourmethod with strong baselines, starting with a qualitative comparison that shows improved subject-consistency and better motion-alignment. We then conduct an ablation study to examine how self-attention query (Q) tokens affect motion and identity, highlighting the contributions of the components in our method.
Finally, quantitative evaluation follows, including a large-scale user study, which demonstrates that users typically favor our results.

\subsection{Evaluation baselines}
We compare our method to several baselines:
\textbf{(1) VideoCrafter2:} A baseline ``vanilla'' text-to-video model, without adaptations. VideoCrafter2 is a public SoTA video model \citep{vbench}. 
\textbf{(2) Tokenflow-Encoder:}  A combination of TokenFlow \cite{tokenflow2023} with IP-Adapter, a Personalization-Based Encoder \citep{ye2023ip}. We personalize TokenFlow by conditioning the IP-Adapter on the first frame of one video generated by the vanilla model. For IP-Adapter we use a high-scale hyper-parameter to push the model toward stronger consistency. 
\textbf{(3) ConsiS Im2Vid:} A combination of SoTA \textit{image}-consistency approach \citep{tewel2024training}, with a subsequent Image-to-Video variant of VideoCrafter. First, we generate a set of consistent \textit{reference} images. Then, we use them as inputs to an Image-to-Video model. We chose VideoCrafter, as it is a public image-to-video model that has an overall quality equivalent to that of the text-to-video VideoCrafter2 model according to the VBench benchmark \cite{vbench}.
\textbf{(4) VSTAR:} A method for generating a long video with dynamic evolution \citep{VSTAR}. We directly provide the multiple prompts and sample 16 frames per prompt, then splitting the result into individual shots.
\textbf{(5) Turbo-V2:} A recent state-of-the-art text-to-video model \cite{turbo_v2} that we use to demonstrate our method's adaptability to other architectures.

\subsection{Qualitative Results}
To visually assess both multi-shot consistency and motion quality in videos, we present two elements per video shot: the initial frame for comparing consistency between shots, and a spatiotemporal slice of the space-time volume, termed "y--t slice" \cite{pmlr-v235-cohen24a}, to visualize motion quality. The selected column for the y--t slice is marked by a yellow line. Typically, we choose the column with the maximum variance in the vanilla-generated video shot. Occasionally, we manually select the y--t column to highlight specific motion characteristics. For ConsiS Im2Vid, the max-variance column is chosen independently, as it does not directly correspond to the vanilla model.

In \cref{fig_qual} and \cref{fig_qual_extra}, we showcase qualitative comparisons between our approach, the vanilla model, and the baselines. Our method demonstrates the ability to alter subject identities consistently across shots, while guiding them towards a unified appearance. This consistency is evident when comparing image frames from different shots. Additionally, an examination of the y--t motion slices reveals that our approach successfully adheres to the motion guided by the vanilla model.

The Tokenflow-Encoder baseline preserves the original motion from vanilla models while primarily affecting the color palette and color style of objects and scenes in videos. However, its impact on the identity of the subject is less pronounced than our approach. Additionally, the combination with a high-scaled IP-Adapter often degrades video quality, causing blurring and color artifacts. See the bird example in \cref{fig_qual} (3rd row) and the boy in \cref{fig_qual_extra} (3rd row).

The ConsiS Im2Vid baseline maintains consistency in its \textit{reference} images. However, the subsequent image-to-video model introduces certain limitations. It lacks awareness of the consistency requirement and the capability to maintain it, causing the subject identity to vary between video shots. Although consistency is maintained within each shot, overall consistency with the reference image is compromised, as seen in the bird example in Figure 1 (4th row). Additionally, the image-to-video model fails to account for the action specified in the text prompt. This results in either minimal motion or movement that does not align with the prompt, as the model relies solely on the conditioning image and cannot effectively utilize the textual information. See the limited motion in the y--t slices in \cref{fig_qual} (4th row) and the corresponding videos in the supplemental material.

VSTAR (\cref{fig_qual_extra}, Appendix) produces large motion dynamics, but struggles with prompt control, often resulting in entire videos misaligning with text descriptions. As it maintains consistency through continuous video generation, it better suits scene transitions than independent shots.

When applied to Turbo-V2 (\cref{fig_turbo_v2}), our method enables subject consistency while leveraging Turbo-V2's enhanced motion capabilities.

\begin{figure}[h]
    \centering
    \makebox[\textwidth][c]{%
        \hspace{-.1cm}%
        \begin{minipage}[t]{8cm}
            \includegraphics[width=\linewidth, trim={0.1cm 0cm 0cm 0cm},clip]{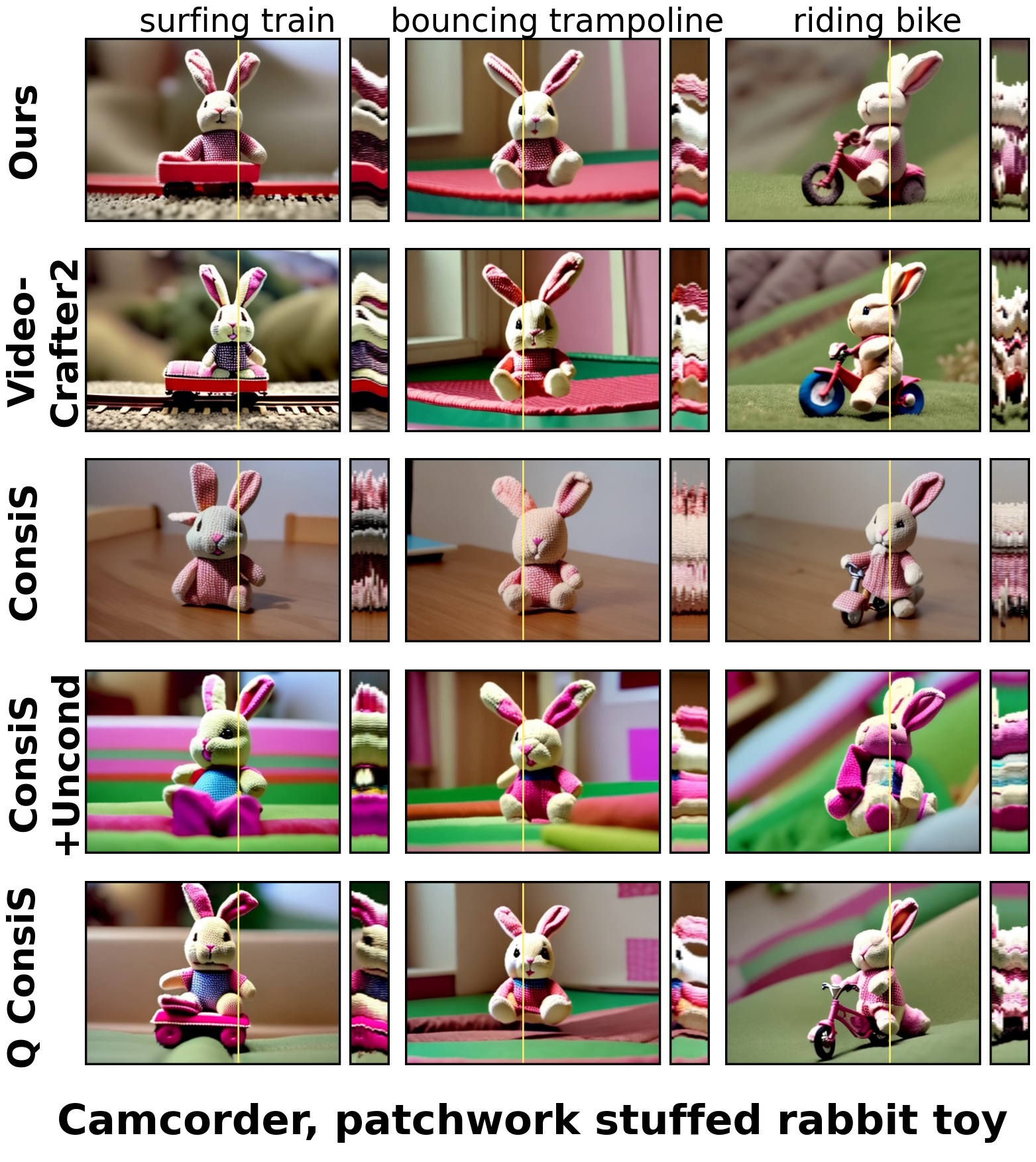}
        \end{minipage}%
        \hspace{0.3cm}%
        \begin{minipage}[t]{7.45cm}
            \includegraphics[width=\linewidth, trim={3.07cm 0.1cm 0cm 0cm},clip]{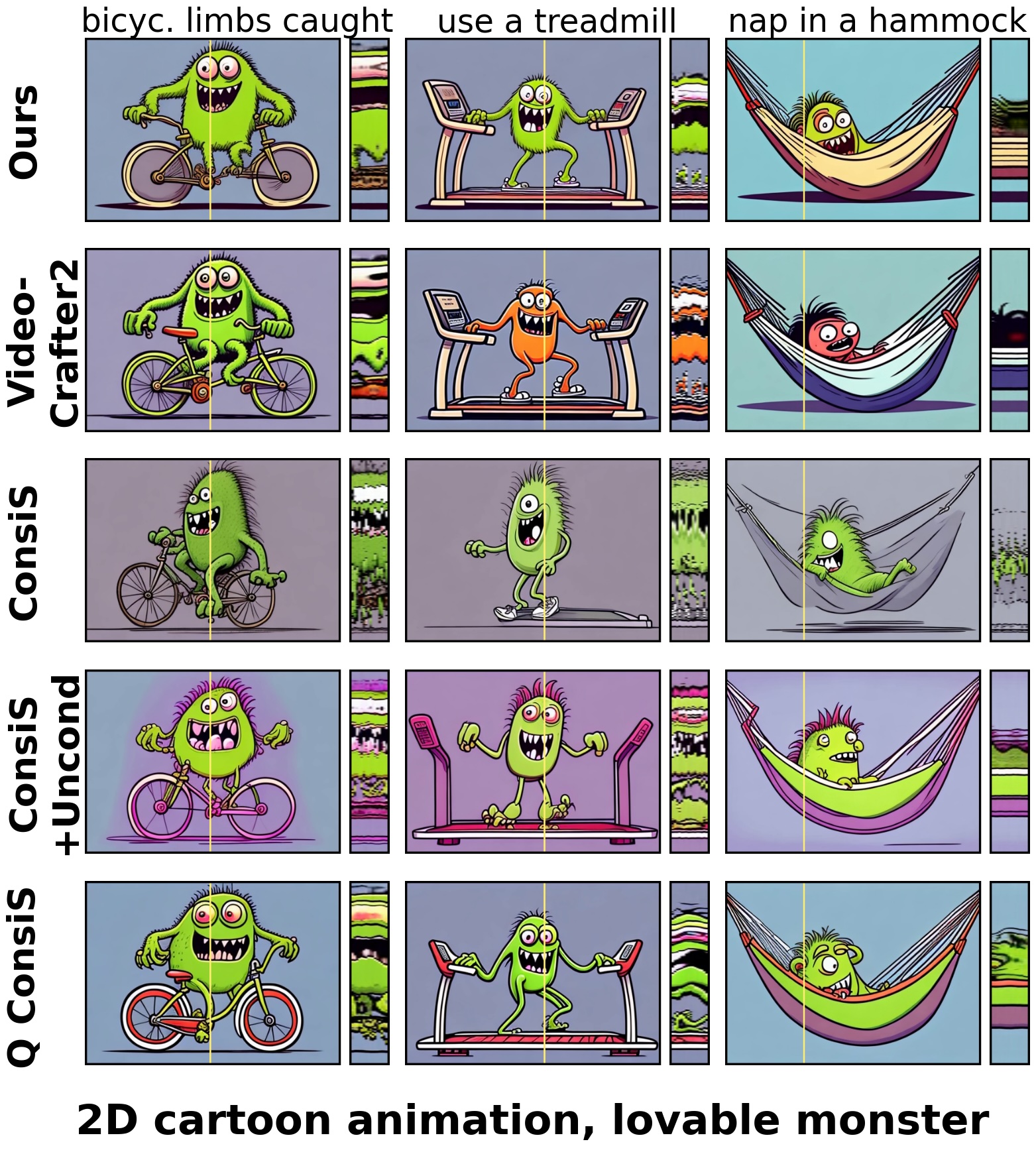}
        \end{minipage}%
    }
    \caption{\textbf{Ablation Study on ConsiStory Components for Video Generation.}  \textit{``Ours" (top row)} demonstrates improved motion richness and identity preservation. \textit{VideoCrafter2 (second row)} shows diverse motion but inconsistent characters. \textit{``ConsiS" (third row)}, a naive ConsiStory implementation, shows impaired identity and motion artifacts. \textit{``ConsiS +Uncond" (fourth row)} adds feature injection to unconditional denoising, resolving motion artifacts but reducing motion magnitude and compromising identity. \textit{``Q ConsiS" (fifth row)} couples each frame with a single frame in an anchor video, allowing some natural motion, although partially synchronized, with improved identity. Our method achieves the best balance of motion quality and identity.}

    \label{fig_ablation_cs}
\end{figure}

\textbf{Adapting ConsiStory for Video Generation.}
Next, we demonstrate the challenges of adapting the image-based ConsiStory algorithm \cite{tewel2024training} to video generation. \cref{fig_ablation_cs} (3rd row ``ConsiS'') shows a naive implementation of ConsiStory with subject-driven extended attention coupled across all frames in each video shot, using subject mask dropout and omitting feature injections to the unconditioned diffusion pass. At each step, it also employs queries influenced by the consistency-preserving mechanism of previous steps, rather than queries from an independent vanilla denoising process.
This results in impaired identity consistency, strong motion artifacts, and unnatural motion flow of different body parts for both the rabbit and monster examples.
Adding feature injection to the unconditional feature denoising (4th row ``ConsiS +Uncond'') resolves motion artifacts but largely reduces motion magnitude (\eg body postures are mostly frozen), and compromises identity. Next, coupling each frame in a shot with a single frame in an anchor video and avoiding SDSA dropout (5th row ``Q ConsiS'') allows for subtle natural motion, although it remains partially synchronized. It also improves identity preservation to some degree. Unlike ConsiStory, SDSA dropout in videos hurts identity without significantly improving motion.
Finally, our method (1st row - ``Ours'') employs a novel Q intervention mechanism. It achieves richer motion with better identity and adherence to the original motion of the vanilla model.

\subsection{Quantitative evaluation}
We conducted a quantitative analysis using automated metrics and a user study, based on a benchmark dataset that we created to assess set-consistency in video generation.

\textbf{Benchmark Dataset:}
We constructed a benchmark dataset of 30 video sets, each containing 5 video-shots with shared subjects but varying prompts. See further details in Appendix \ref{sec_benchmark_dataset}.

\textbf{Evaluation Protocol:}
To avoid overfitting, we conducted all development and parameter tuning on a separate collection of 16 distinct subject-prompt sets. The test set was used exclusively for final evaluations, without any component development or hyperparameter tuning. 

\textbf{Evaluation Metrics:}
Our evaluation approach builds on previous work in image consistency and personalization \cite{tewel2024training,gal2022textual,ruiz2022dreambooth}, focusing on multi-shot set-consistency and motion dynamics. For \textbf{set-consistency}, we measure average pairwise DINO feature similarity \cite{caron2021emerging,vbench} across all frames in a set, excluding pairs within the same video shot. We isolate the subject by masking out the background \cite{fu2023dreamsim} before extracting each frame's features, using ClipSEG \cite{luddecke2021prompt} with a dynamic threshold determined by ``Otsu's method'' \cite{otsu}.
For \textbf{motion dynamics}, we evaluate all $150$ generated videos using VBench's "Dynamic Degree" metric \cite{vbench}, which classifies the significance of video motion by measuring RAFT-based optical flow intensity. We focused on motion dynamics over text prompt alignment due to two challenges: actions are often visible even in videos with minimal motion, making it difficult for temporal CLIP-like models \cite{wang2024viclip} to distinguish between our method and baselines; also, sharing seeds across baselines lead to similar visual structures, with main differences in motion quality. We include text-similarity metrics in Table \ref{table_metrics} (Appendix), measuring temporal CLIP similarity between each video shot and its prompt.

\begin{figure}[htbp]
    \centering
    \begin{minipage}{0.5\textwidth}
        \includegraphics[width=\textwidth, trim={0.cm 0cm 0cm 0cm},clip]{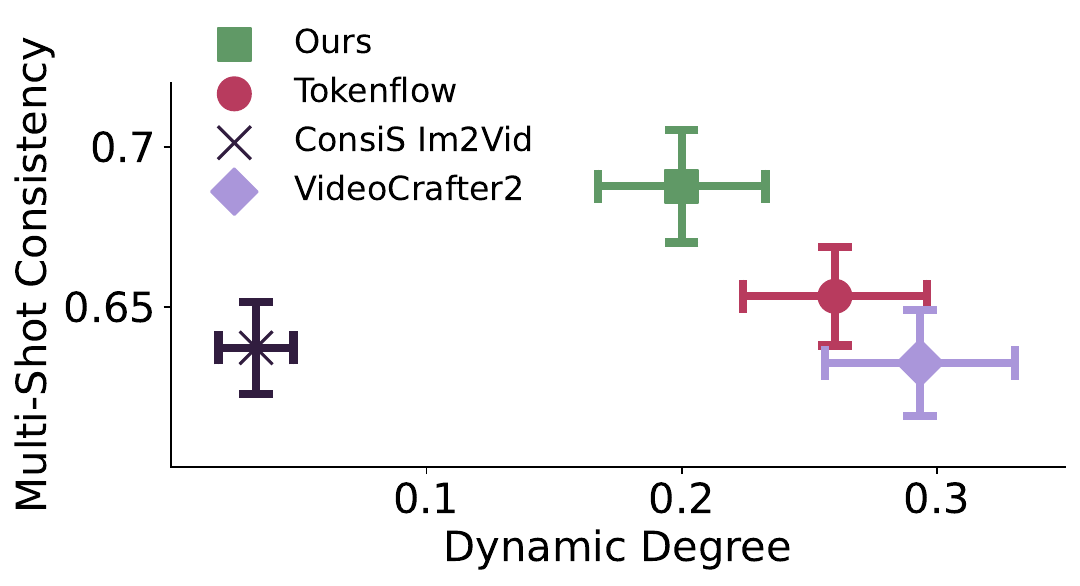}
    \end{minipage}%
    \hspace{0.2cm}
    \begin{minipage}{0.4\textwidth}
        \caption{\textbf{Quantitative Evaluation of Set Consistency and Motion Dynamics:} Our approach achieves highest set consistency score while maintaining competitive motion dynamics. Error bars indicate standard error of the mean.}
        \label{fig_quant}
    \end{minipage}

\end{figure}

\textbf{Results:} \cref{fig_quant} show our approach  enhances multi-shot set consistency, while sacrificing  motion magnitude compared to vanilla VideoCrafter2. Tokenflow-Encoder baseline shows consistency improvement and slight motion decrease. ConsiS-Im2Vid baseline's performance aligns with qualitative analysis, showing low motion scores.
A comparison of all baselines, including VSTAR and Turbo-V2, is presented in Table \ref{table_metrics} (Appendix). VSTAR struggles with prompt control (19.8 vs 27.7 for ours), while achieving the highest consistency and motion dynamics.  When combined with Turbo-V2, our method improves multi-shot consistency while maintaining high motion quality: The dynamic degree improves threefold, from 20 to 62, while keeping the same level of text alignment.

These quantitative results offer insights into trade-offs between our approach and baselines, but cannot fully capture user-perceived quality or alignment of generated motions with text prompts. Therefore, we conducted a comprehensive user preference study using two and three-alternative forced-choice format, focusing on two key aspects: set-consistency and text-motion alignment. For set-consistency, users selected the better set from two sets of 5 videos each depicting the subject. For text-motion alignment, users chose the video best matching the action described in the prompt from a pair of videos. To distinguish between degraded motions and those largely unchanged, users could also indicate if motion quality was equivalent in both videos.
We used the same test benchmark as the automated metric study, collecting $5$ repetitions per question for set-consistency and $3$ repetitions for text-motion alignment, totaling $1800$ responses.

\begin{figure}[h]
    \setlength{\abovecaptionskip}{4pt}
    \setlength{\belowcaptionskip}{0pt}
    \centering
    \includegraphics[width=0.42\columnwidth, trim={0.cm 0cm 8.7cm 0cm},clip]{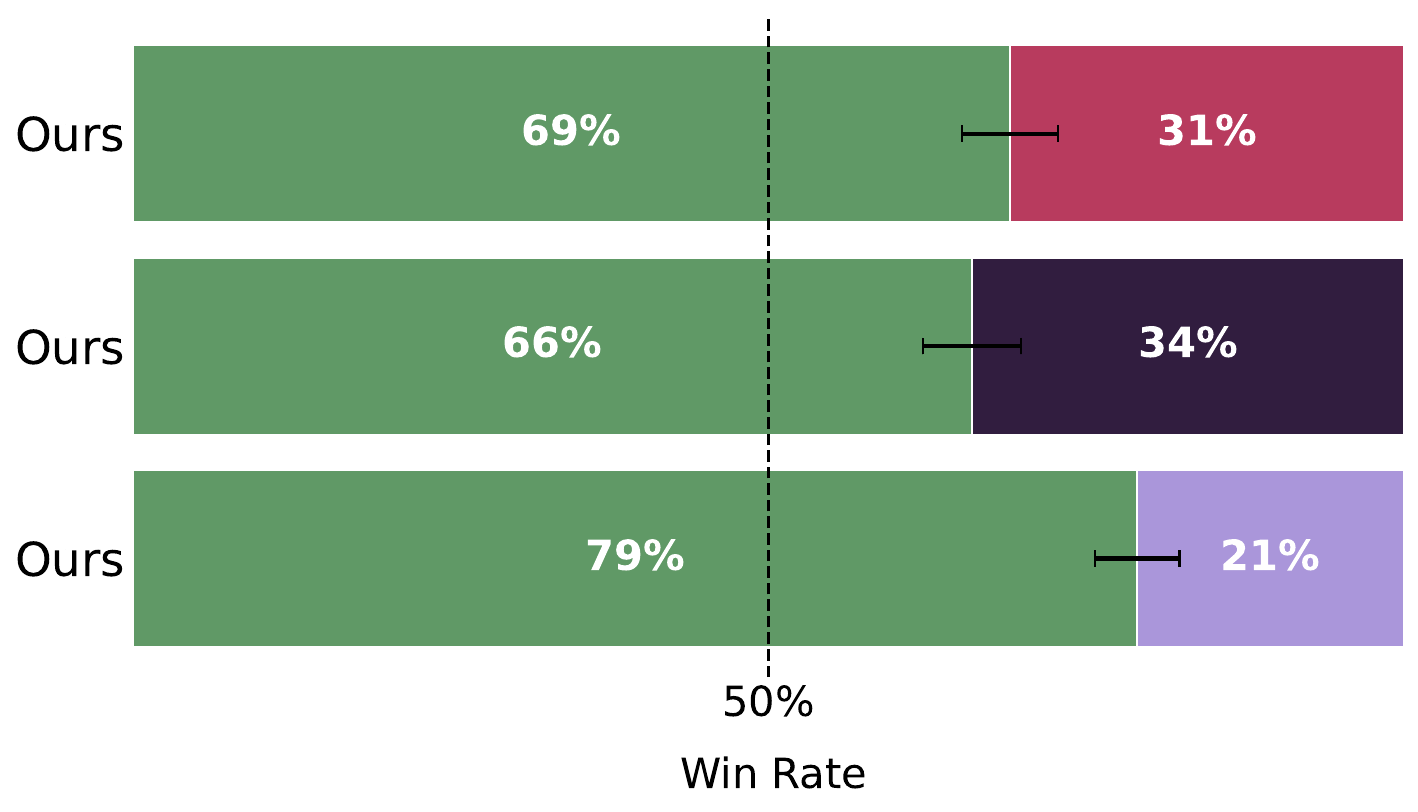} %
    \hspace{0.3cm}
    \includegraphics[width=0.535\columnwidth, trim={2.1cm 0cm 0cm 0cm},clip]{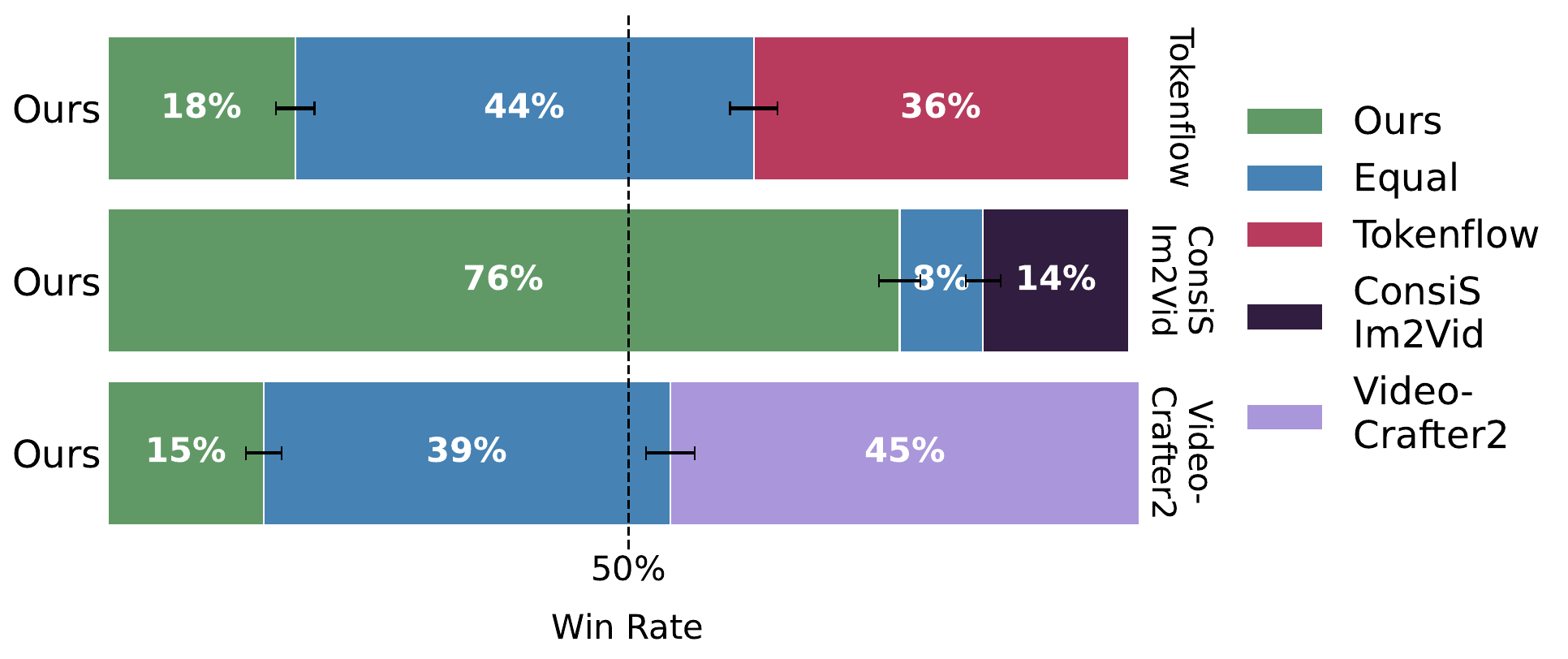} %
    \caption{\textbf{User Study:} \textit{(left)} We measure user preferences for set consistency  and \textit{(right)} how well the generated motion matches the text prompt . Our approach achieves the superior set consistency score while maintaining competitive text-motion alignment. Notably, 55\% of our generated motions were judged to be of similar or better quality compared to the vanilla model. Error bars are S.E.M.}
    
    \label{fig_userstudy}
    \vspace{10pt}  
\end{figure}

The user-study results in \cref{fig_userstudy}, reveal that \ourmethod outperforms the baselines in set consistency. For motion quality, $55\%$ of users rated the generated motions as similar or superior to those of the vanilla model. The ConsiS-Img2Vid baseline's motion quality was consistent with our earlier findings, showing lower motion quality. However, it achieved the highest set consistency among the baselines, winning in $34\%$ of the generated sets compared to our approach.

\subsection{Additional Results}
\label{supp_qual_extra}

\begin{figure}[h]
    \centering    
    \makebox[\textwidth][c]{%
        \hspace{-.1cm}%
        \begin{minipage}[t]{8cm}
            \includegraphics[width=\linewidth, trim={0.1cm 0cm 0cm 0cm},clip]{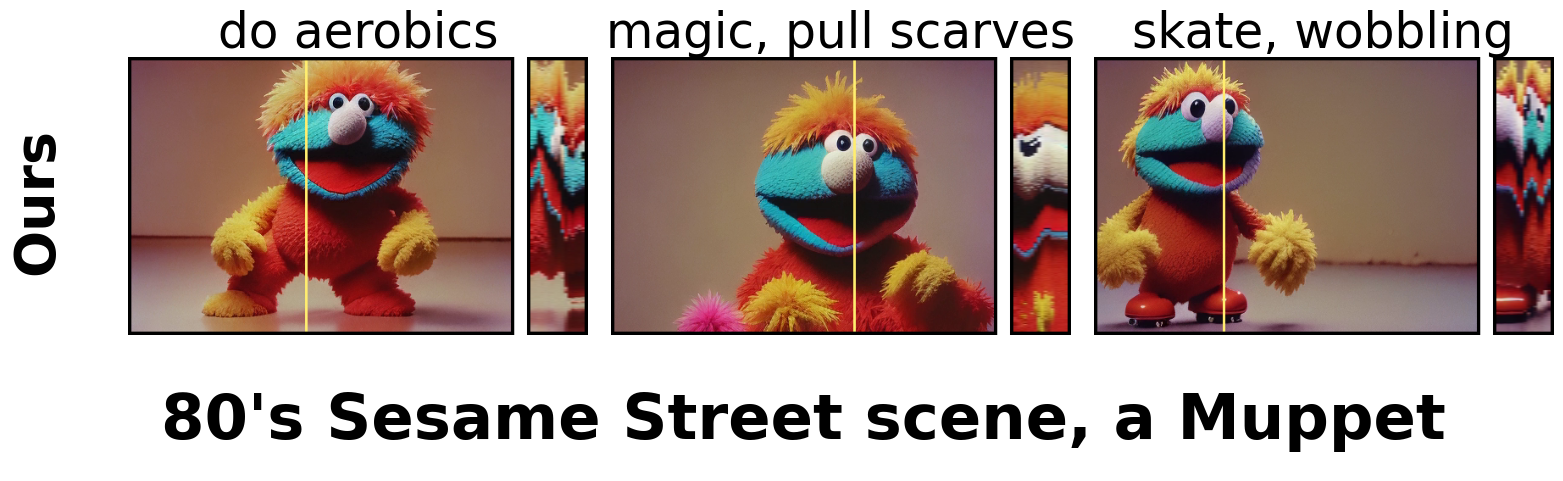}
        \end{minipage}%
        \hspace{0.3cm}%
        \begin{minipage}[t]{7.45cm}
            \includegraphics[width=\linewidth, trim={2.3cm 0cm 0cm 0cm},clip]{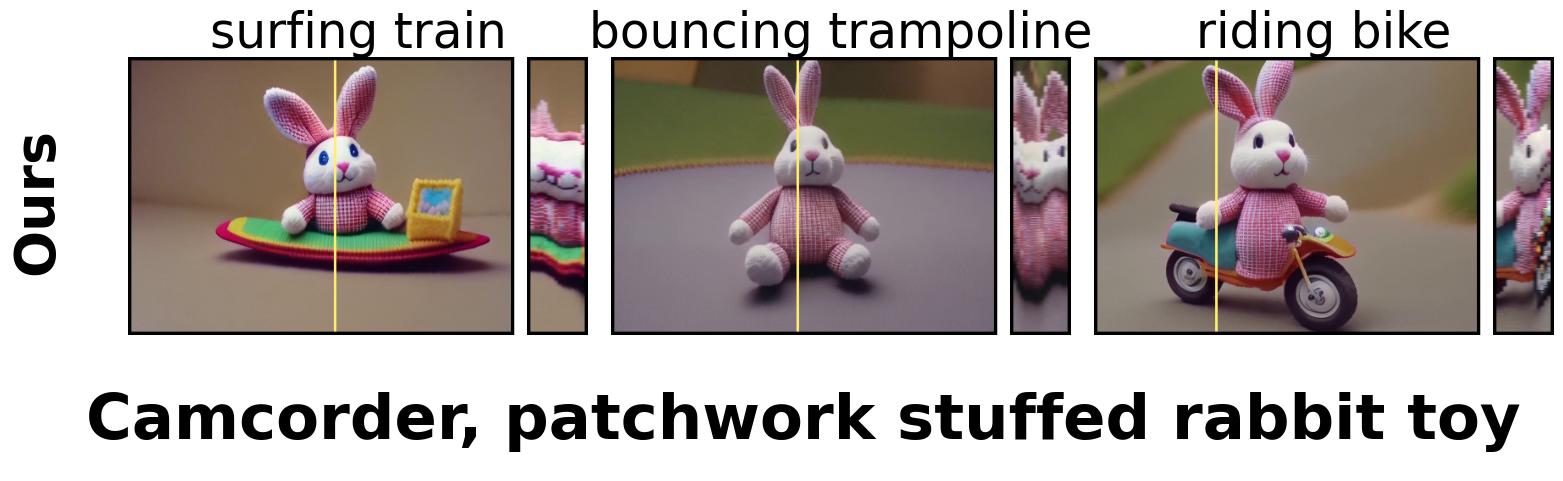}
        \end{minipage}%
        }
    \caption{\textbf{T2V-Turbo-V2:} 
    \ourmethod can be applied to T2V-Turbo-V2 \cite{turbo_v2}, a recent state-of-the-art video model, that exhibits significantly better motion.}
    \label{fig_turbo_v2}
\end{figure}

\cref{fig_turbo_v2} illustrates the adaptability of our method when applied to the state-of-the-art T2V-Turbo-V2 model \citep{turbo_v2}. The results show enhanced motion quality while maintaining subject consistency, demonstrating that our approach can effectively improve even the most recent video generation models.

\begin{figure}[htbp]
    \centering
    \makebox[\textwidth][c]{%
        \hspace{-.1cm}%
        \begin{minipage}[t]{6.55cm}
            \includegraphics[width=\linewidth, trim={0.1cm 0cm 0cm 0cm},clip]{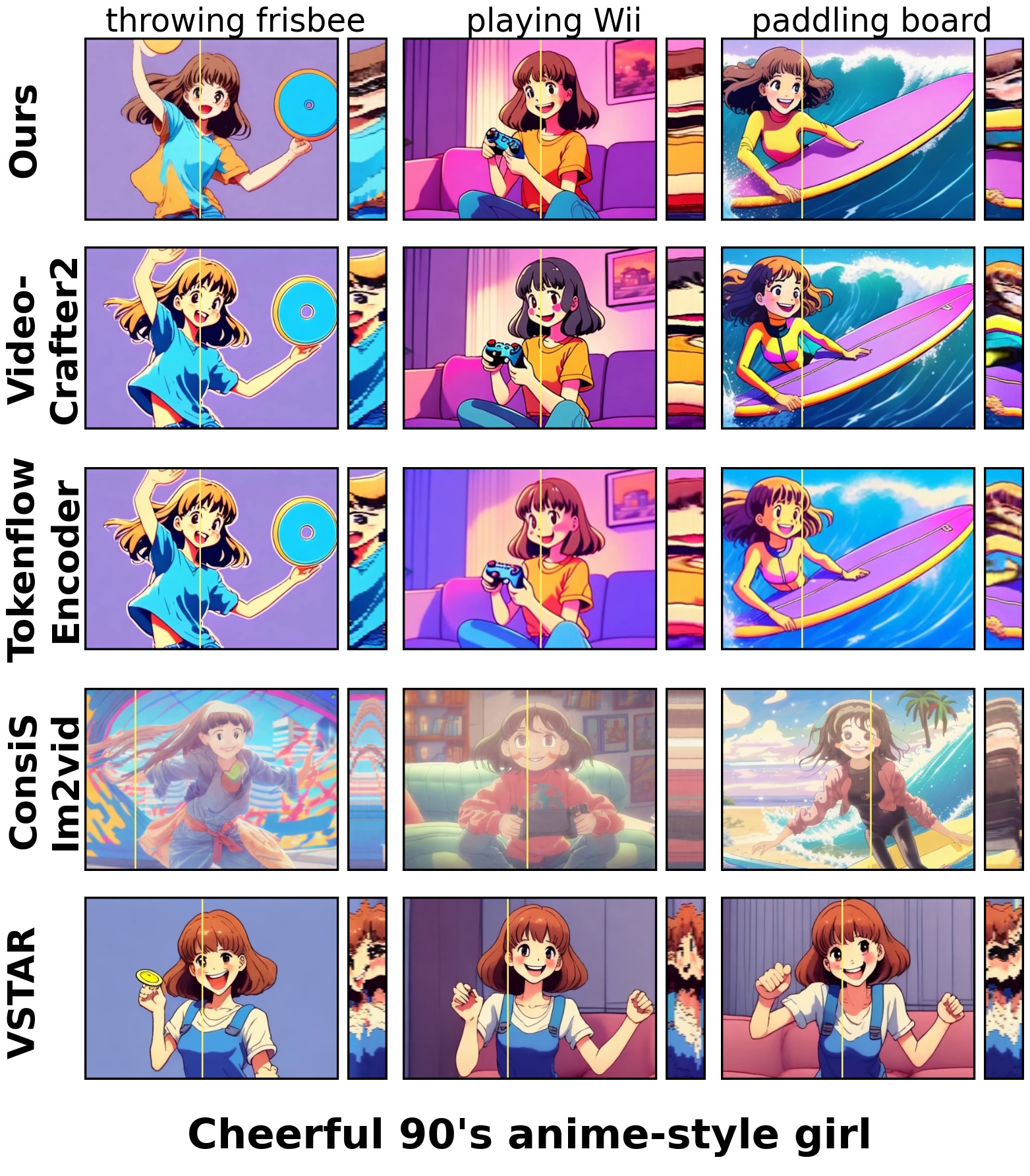}
        \end{minipage}%
        \hspace{0.3cm}%
        \begin{minipage}[t]{6.0cm}
            \includegraphics[width=\linewidth, trim={3.07cm 0cm 0cm 0cm},clip]{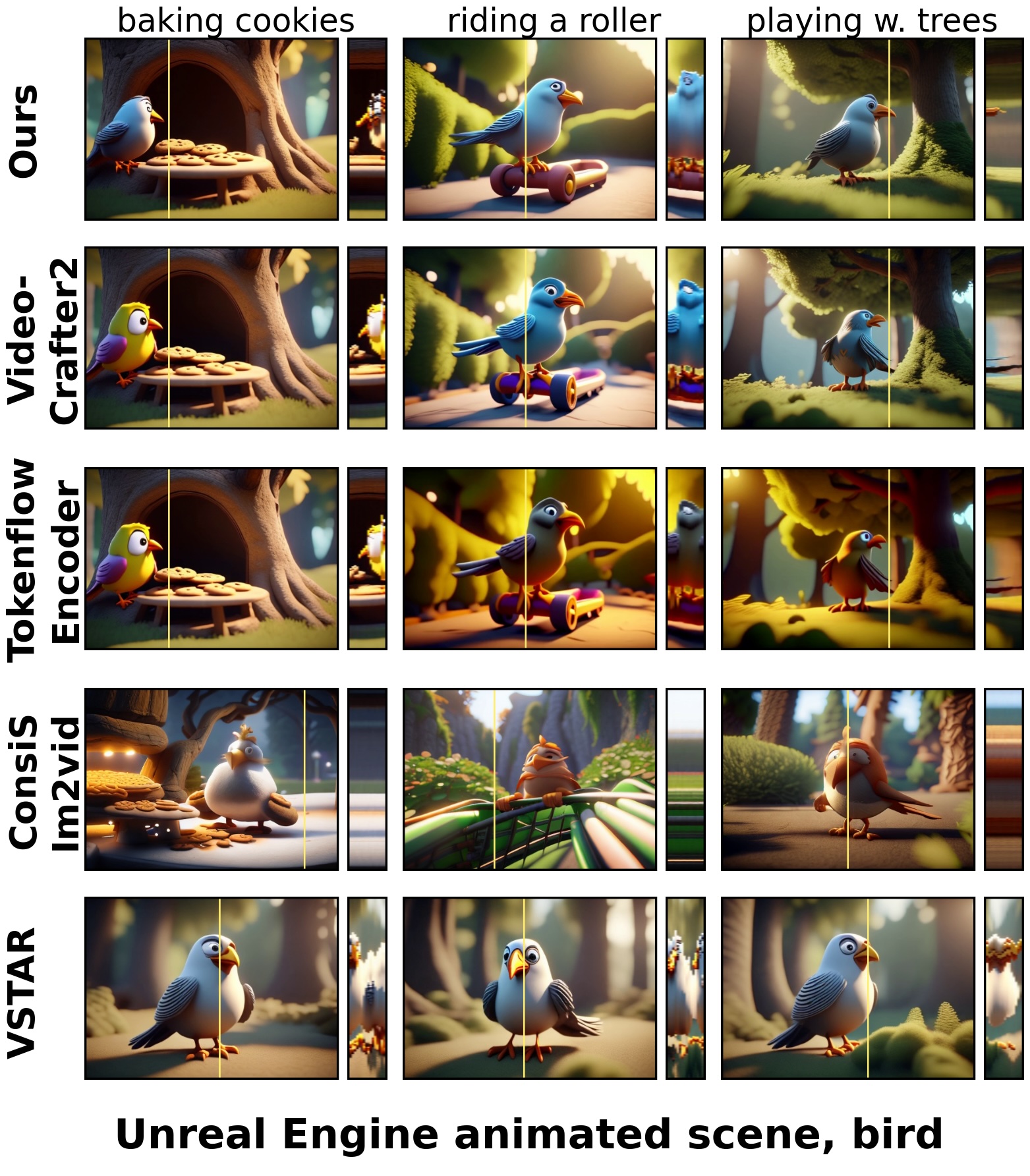}
        \end{minipage}%
        }
    \makebox[\textwidth][c]{%
        \hspace{-.1cm}%
        \begin{minipage}[t]{6.55cm}
            \includegraphics[width=\linewidth, trim={0.1cm 0cm 0cm 0cm},clip]{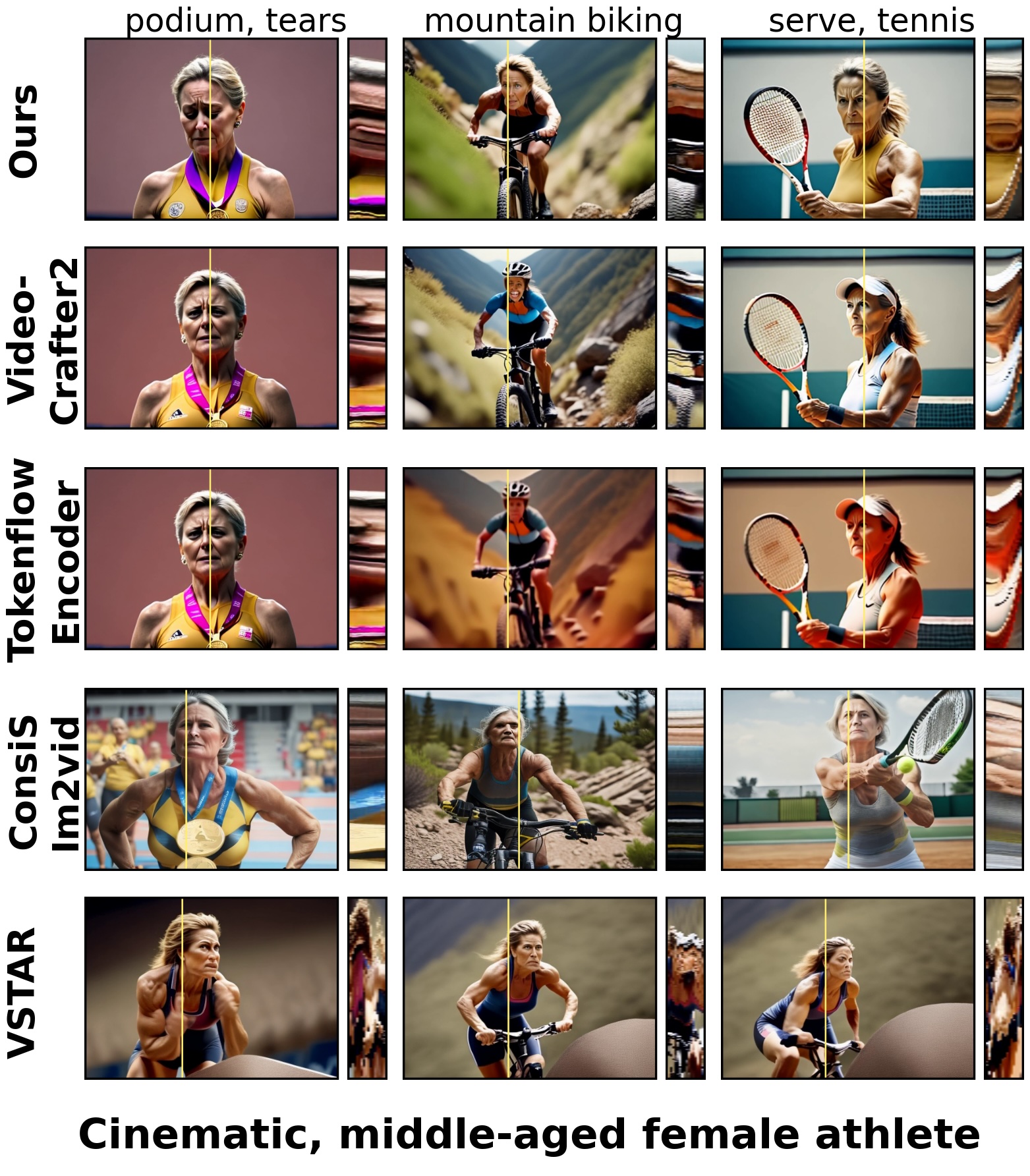}
        \end{minipage}%
        \hspace{0.3cm}%
        \begin{minipage}[t]{6.0cm}
            \includegraphics[width=\linewidth, trim={3.07cm 0cm 0cm 0cm},clip]{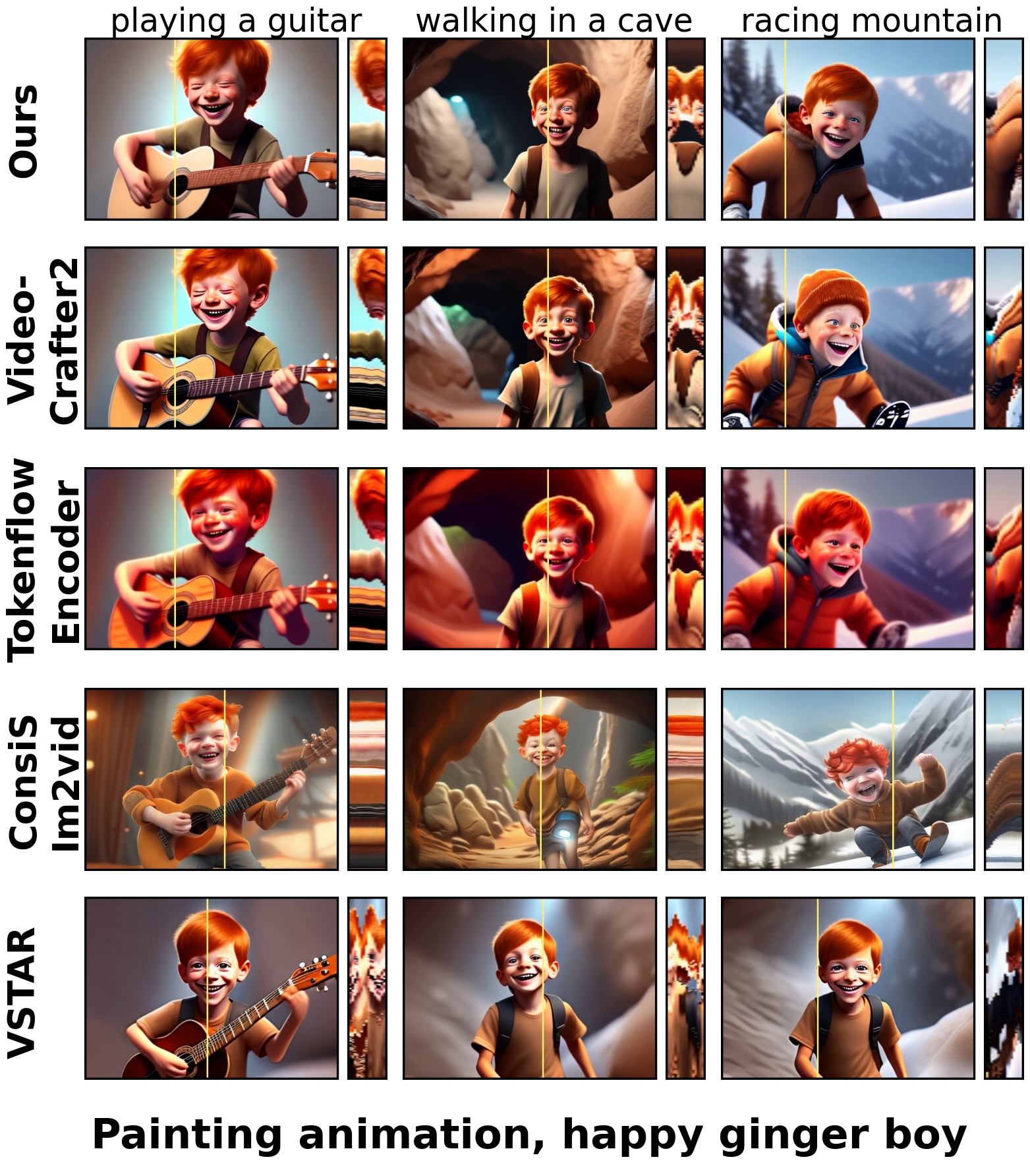}
        \end{minipage}%
        }
    \makebox[\textwidth][c]{%
        \hspace{-.1cm}%
        \begin{minipage}[t]{6.55cm}
            \includegraphics[width=\linewidth, trim={0.1cm 0cm 0cm 0cm},clip]{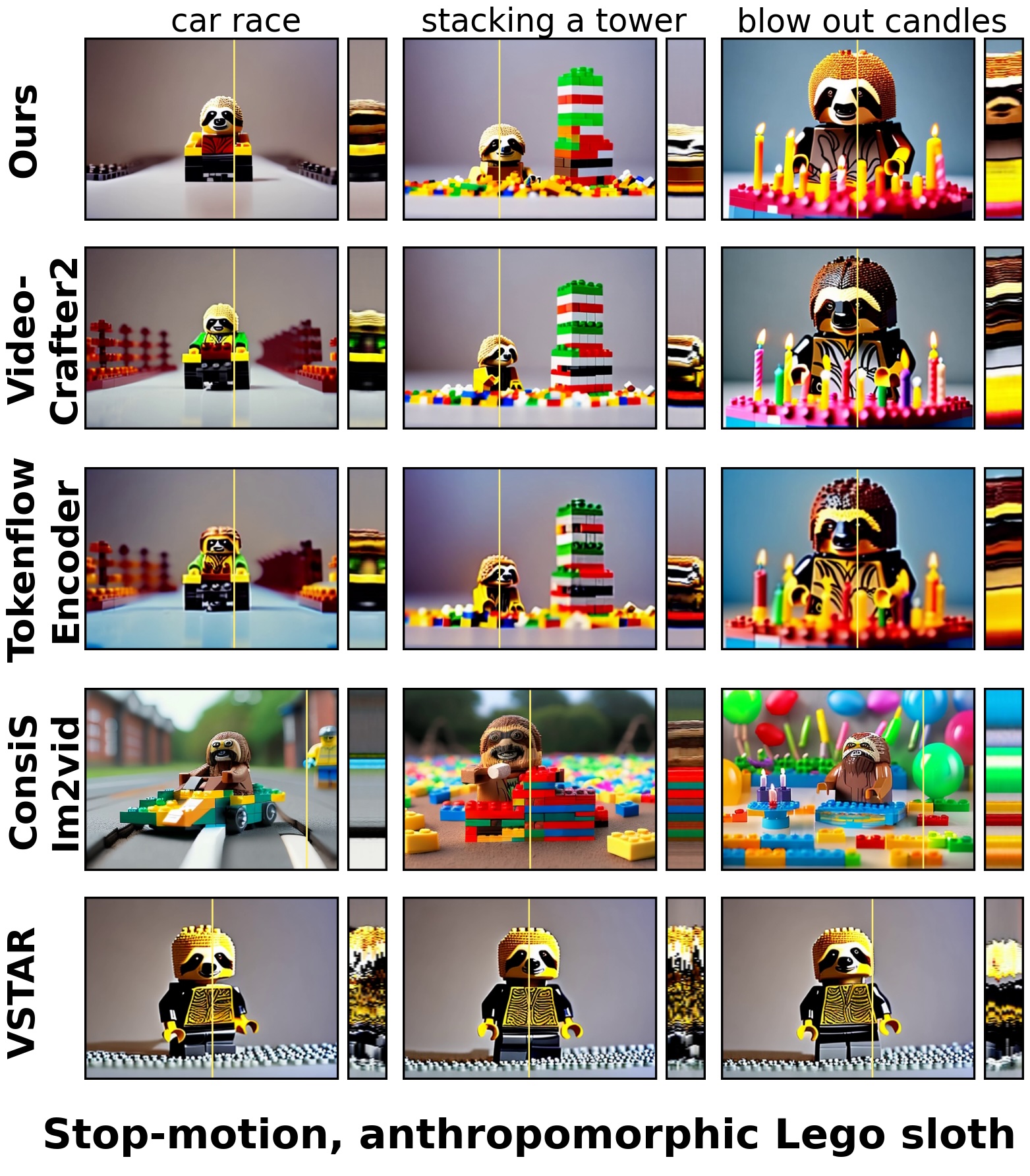}
        \end{minipage}%
        \hspace{0.3cm}%
        \begin{minipage}[t]{6.0cm}
            \includegraphics[width=\linewidth, trim={3.07cm 0cm 0cm 0cm},clip]{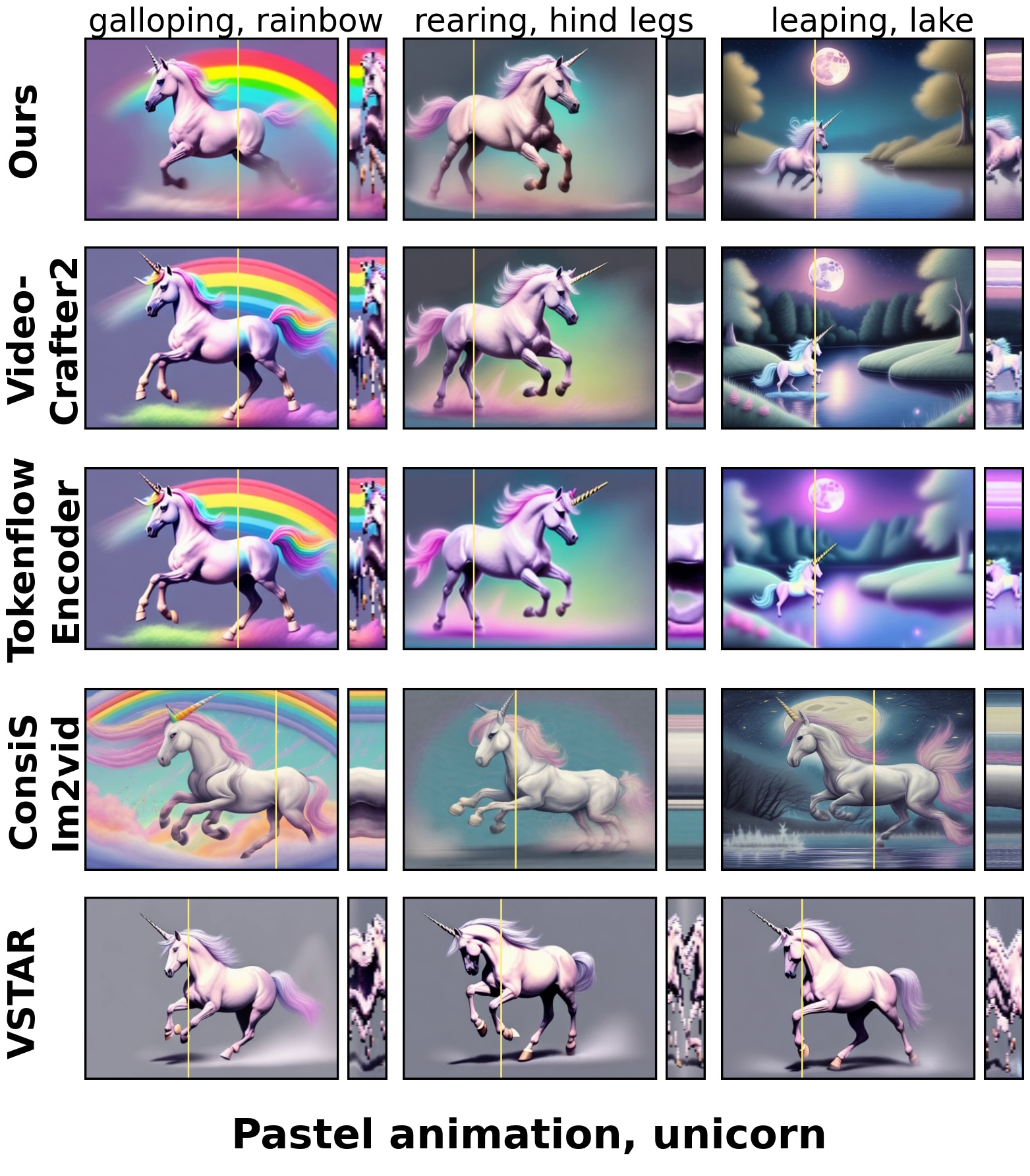}
        \end{minipage}%
    }    
    \vspace{-20pt}
    \caption{\textbf{Additional Qualitative Comparisons , including VSTAR:}  Our method generates consistent subjects while preserving diverse and natural motions across scenarios.}
    \label{fig_qual_extra}
\end{figure}

\cref{fig_qual_extra}, provides additional qualitative comparisions to \cref{fig_qual}, and also includes qualitative comparison with VSTAR baseline \citep{VSTAR}.

In Table \ref{table_metrics} we present a comprehensive quantitative comparison across different models using three key metrics. Our method, when combined with both VideoCrafter2 and Turbo-V2, shows improved Multi-Shot Consistency scores (68.8 and 67.3 respectively) compared to their baseline versions (63.2 and 63.3), while maintaining comparable Text Similarity and Dynamic Degree measurements. This indicates that our approach successfully enhances subject consistency without significantly compromising other important aspects of video generation. In the reported metrics, we also include a ``Subject-Consistency'' metric, introduced by VBench \citep{vbench}. This metric measures the similarity between frames within the same video shot using DINO (see Table 1 in the Appendix).

\begin{table}[htbp]{
    \begin{sc}
    \scalebox{0.9}{
    \renewcommand{\arraystretch}{1.}
        \begin{tabular}{lcccc}
        \toprule
         & \makecell{Multi-Shot \\ Consistency} & \makecell{Text \\ Similarity} & \makecell{Dynamic \\ Degree} & \makecell{Subject \\ Consistency} \\
        \midrule
        ConsiS Im2Vid & 63.7 $\pm$ 1.4 & 27.3 $\pm$ 0.5 & 3.3 $\pm$ 1.5 & 99.1 $\pm$ 0.1 \\
        VSTAR & 83.9 $\pm$ 1.6 & 19.8 $\pm$ 0.4 & 90.7 $\pm$ 2.4 & 92.6 $\pm$ 0.3 \\
        Tokenflow & 65.3 $\pm$ 1.5 & 27.9 $\pm$ 0.4 & 26.0 $\pm$ 3.6 & 97.7 $\pm$ 0.2 \\
        VideoCrafter2 & 63.2 $\pm$ 1.7 & 28.7 $\pm$ 0.4 & 29.3 $\pm$ 3.7 & 97.3 $\pm$ 0.2 \\
        Ours + VideoCrafter2 & 68.8 $\pm$ 1.8 & 27.7 $\pm$ 0.4 & 20.0 $\pm$ 3.3 & 97.7 $\pm$ 0.2 \\
        \midrule
        Turbo-V2 & 63.3 $\pm$ 1.7 & 28.6 $\pm$ 0.4 & 63.3 $\pm$ 3.9 & 96.2 $\pm$ 0.2 \\
        Ours + Turbo-V2 & 67.3 $\pm$ 2.1 & 27.4 $\pm$ 0.4 & 62.0 $\pm$ 4.0 & 96.8 $\pm$ 0.2 \\
        \bottomrule
        \end{tabular}
    }
    \end{sc}
    
    \vspace{5pt}
    \caption{\textbf{Quantitative Evaluation Metrics.} Comparison of different models across three metrics: Multi-Shot Consistency, Text Similarity, and Dynamic Degree. Values are reported as mean $\pm$ standard error of the mean (S.E.M).}
    \label{table_metrics}
    }
\end{table}
\subsection{Framewise Subject-Driven Self-Attention - Implementation Details}
\label{fSDSA_formal}
This section provides a detailed explanation of our proposed Framewise-SDSA mechanism.

\paragraph{Improved Subject Localization.}
In video generation, subject localization becomes particularly challenging during early denoising steps, where the noise is most prominent. aggregation method proposed in ConsiStory (Sec.~\ref{sec:consistory_details}) proved insufficient in this context, particularly during the earliest denoising steps, leading to unreliable masks both in terms of accuracy and false positive localization.

To address this, we propose using the estimated clean image $\hat{x_0}$ for subject localization instead of relying on internal network activations. At each denoising step $t$, we estimate $\hat{x_0}$ from the noisy latent $x$ using: $\hat{x_0} = \big({x - \sqrt{1 - \alpha_t} \cdot e_t} \big) /{\sqrt{\alpha_t}}$, where $e_t$ is the estimated noise, and $\alpha_t$ is the schedule parameter \cite{song2020denoising}. We then apply a zero-shot segmentation approach~\citep{luddecke2021prompt} to localize the subject in the estimated image, followed by Otsu's method~\citep{otsu} to dynamically threshold the mask. This approach produces reliable subject masks from the earliest denoising steps and throughout the generation process.

\paragraph{Maintaining Motion Fluidity.}
Our experiments revealed that a direct application of SDSA -- attending to all frames across all videos simultaneously -- can lead to visual artifacts and frozen motion. We discovered that limiting attention to a single corresponding frame in other shots is most effective, as attending to two or more frames negatively impacts motion fluidity and introduces visual artifacts. Specifically, we propose a framewise attention scheme. Instead of attending to all frames across all video shots, frames with matching temporal indices across shots attend only to each other. This prevents visual artifacts and frozen motion, which occur when attending to multiple frames simultaneously and strikes a balance between subject consistency and natural motion.

\paragraph{Formal Definition of Framewise-SDSA.}
Let $K_{if},Q_{if},V_{if},M_{if}$ be the keys, queries, values and subject-mask for frame $f$ in video shot $i$. The framewise extended self-attention $A_{if}^{+}$ is defined by:
\begin{equation}
    K^{+}_f=[K_{1,f}\oplus K_{2,f} \oplus \cdots \oplus K_{N,f}] \nonumber
\end{equation}
\begin{equation}
    V^{+}_f=[V_{1,f}\oplus V_{2,f} \oplus \cdots \oplus V_{N.f}] \nonumber
\end{equation}
\begin{equation}
    M_{i,f}^{+}=[M_{1,f}\oplus \cdots \oplus M_{i-1,f} \oplus \mathds{1} \oplus M_{i+1,f} \cdots \oplus M_{N,f}]  \nonumber
\end{equation}
\begin{equation}
    A_{i,f}^{+} = \textit{softmax}\left({Q_i K^{+}_f}/{\sqrt{d_k}} + \log M_{i,f}^{+} \right) \nonumber %
\end{equation}
\begin{equation}
    h_{i,f} = A_{i,f}^{+} \cdot V^{+}_f %
\end{equation}
where $\oplus$ indicates matrix concatenation. We use standard attention masking, which null-out softmax's logits by assigning their scores to $-\infty$ according to the mask. Note that in this step, the Query tokens remain unaltered, and that the concatenated mask $M_{i,f}^{+}$ is set to be an array of $1$'s for patch indices that belong to the $i^{th}$ image itself.

\subsection{Flow-based Q components injection - Formal Definition}
\label{formal_flow_Q}
Let $q_{fxy}\in \mathbb{R}^F$ represent a Q feature from an originally generated video at location $(x, y)$ in frame $f$. We denote by $f_A$ and $f_B$ the indices of the two nearest keyframes, where $f_A \leq f \leq f_B$. The locations of the most similar Q features in frames $f_A$ and $f_B$, denoted by $(x_A, y_A)$ and $(x_B, y_B)$ respectively, are defined as:
\begin{equation}
    (x_A,y_A)=\underset{x_0,y_0}{\text{argmax \ }}{\mathcal{S}_\text{cos}(q_{fxy},q_{f_Ax_0y_0})}
\end{equation}
\begin{equation}
    (x_B,y_B)=\underset{x_0,y_0}{\text{argmax \ }}{\mathcal{S}_\text{cos}(q_{fxy},q_{f_Bx_0y_0})}
\end{equation}
where $\mathcal{S}_\text{cos}(a, b)$ represents the cosine similarity between $a$ and $b$.

We then modify the generated Q feature, denoted by $\hat{q}_{fxy}$, as follows:

\begin{equation}
    \hat{q}_{fxy}=w \hat{q}_{f_Ax_Ay_A}+(1-w)\hat{q}_{f_Bx_By_B}
\end{equation}

where $w = \text{sigmoid}\left( \frac{f_B - f}{f_B - f_A} \right)$. This ensures that $\hat{q}$ maintains the feature flow of the originally generated video, without injecting the actual features from it.

\subsection{Benchmark Dataset Construction:}
\label{sec_benchmark_dataset}
We created a benchmark dataset comprising $30$ video sets, each containing $5$ video-shots depicting a shared subject under different prompts. The evaluation prompts were crafted using the Claude Sonnet 3.5 AI-Agent, following this protocol: each prompt consisted of three parts: (1) a subject description, \eg, \textit{``A girl''} (2) a setting description, \eg, \textit{``paddling out on her surfboard''}, and (3) a style descriptor encompassing both image and motion styles, \eg, \textit{``Anime cartoon animation''} or \textit{``Shaky camcoder footage''}. We instructed the AI-agent to choose actions that are visually striking and could be captured in a split second. Within each set, prompts shared the same subject and style but varied in settings.
To ensure a challenging and representative test set, we selected a subset of 5 prompts per subject, prioritizing those that produced videos with significant motion and subject variability when processed by the vanilla model. Importantly, to ensure fairness, this selection process relied solely on the vanilla model's generations.

\subsection{Q dropout}
\label{suppl_q_dropout}
\begin{figure}[h]
 \centering
 \includegraphics[width=0.5\textwidth, trim={0.cm 4.5cm 0.cm 0cm},clip]{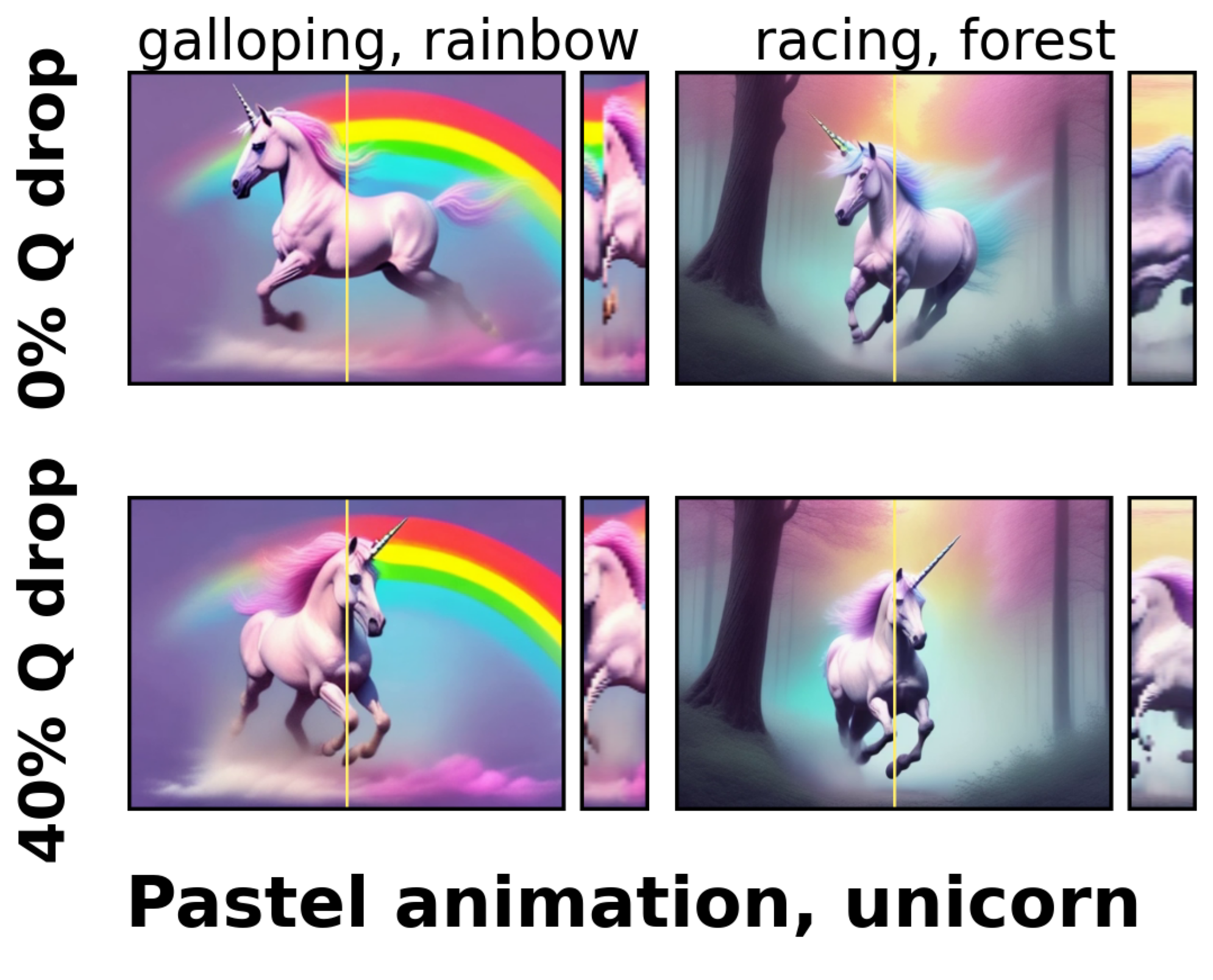}
 \caption{\textbf{Q dropout:} Q injection may hurt identity. Q dropout may trade-off identity for motion. At 0\% the unicorn gallops at both directions. At 40\%, only to the right.}
 \label{fig_q_dropout}
\end{figure}

When Q injection is too strong, it can compromise identity preservation. To address this, we introduce Q dropout, which reduces the strength of Q injection. Unlike SDSA dropout, which hurts identity when trying to improve the image structure, Q dropout sacrifices some visual structural (motion) to enhance identity preservation. This Identity-Motion Trade-off is illustrated in \cref{fig_q_dropout}, where increasing Q dropout improves identity consistency but reduces motion richness.

\subsection{Implementation Details}
\label{supp_implementation_details}

\textbf{Anchor Videos:} Similar to ConsiStory, we utilize two anchor videos that share all features between themselves. Other videos in the batch only observe features derived from these anchors.

\textbf{Scalable Video Batch Processing with Sub-batch Attention:}  To fit large batches of video generation within available GPU memory, we process the self and cross-attention computations in smaller sub-batches. This approach uses an internal loop, and subsequently concatenates results into a single tensor. The operation remains transparent to the network, enabling the generation of larger batches of video shots.

\textbf{Reproducible denoising.} Our pipeline involves three denoising iterations: caching vanilla queries, applying Q injection and Framewise SDSA, and adding refinement feature injection. To ensure consistency across these stages, we maintain identical random generators for both initial noisy latents and the denoising process. This approach guarantees that each part builds upon the previous one, preserving the reliability of our reproducible denoising pipeline.

\textbf{Temporal Parameters:} For Q preservation, we set $t_{pres}$ to 750. Framewise-SDSA is applied for $t \in [550, 950]$. Our refinement feature injection step is employed during $t \in [590, 950]$.

\textbf{Feature Injection:} We apply our refinement feature injection step to the $32 \times 20$ self-attention layers. Other layers either produced visual artifacts or did not significantly affect identity.

\textbf{Denoising Process:} Videos were sampled the default VideoCrafter2 configuration, using 50 DDPM steps with a guidance scale of 12. 

\textbf{T2V-Turbo-V2:} For T2V-Turbo-V2 we adapt our Framewise-SDSA by allowing each frame to attend to both its temporally matching frames across shots and the middle frame of each shot. Other hyper-parameters were kept the same.

\subsection{User Study Protocol}
\label{supp_user_study_protocol}
The following screenshots illustrate the experimental framework used in our user study:

\begin{figure}[ht]
\includegraphics[width=0.9\textwidth, trim={0.cm 0.cm 0.cm 0.cm},clip]{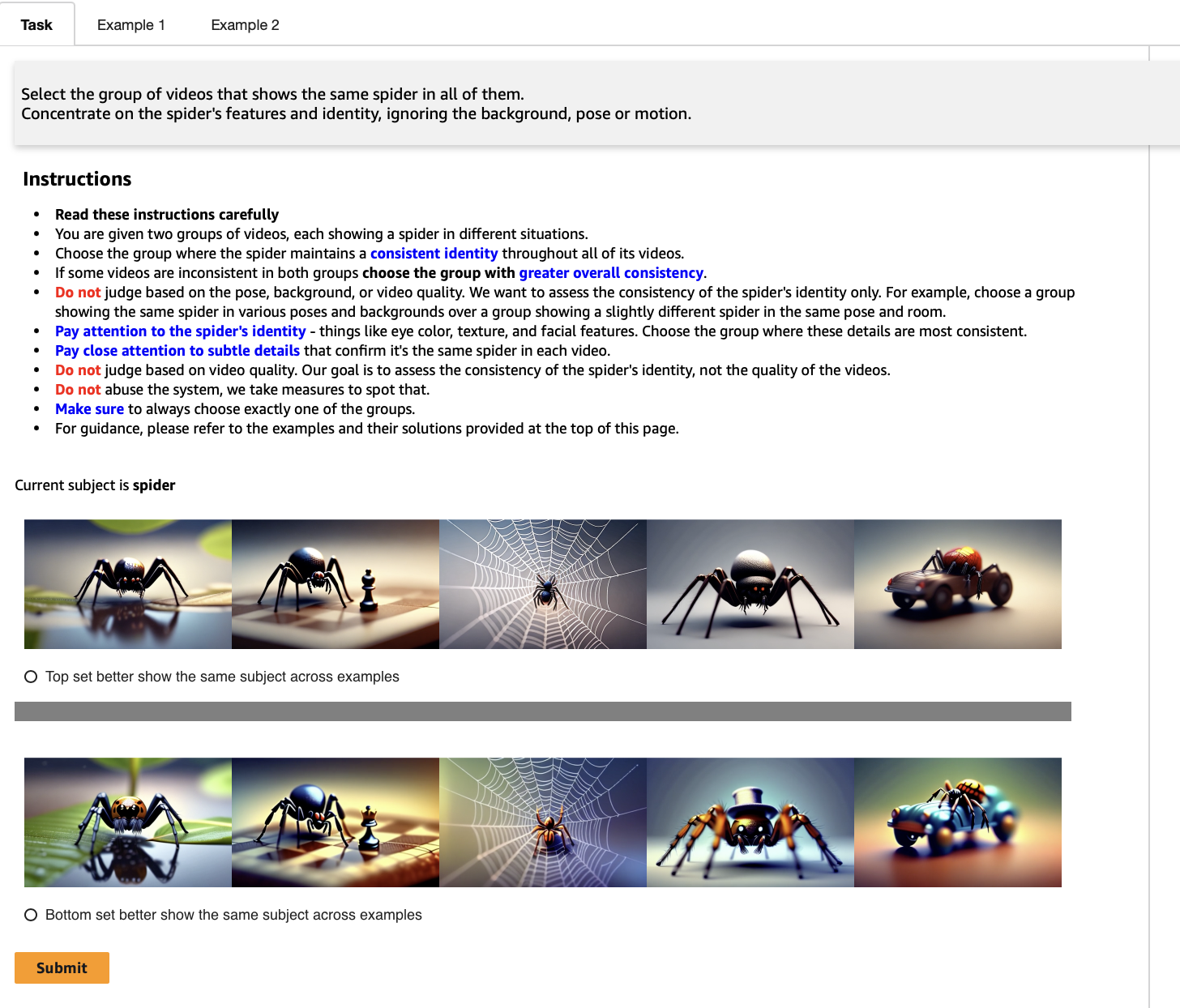} %
\vspace{-15pt}
\caption{One trial of the visual consistency user study.}
\label{fig_amt1_task}

\begin{flushleft}
\includegraphics[width=0.48\textwidth, trim={0.cm 0.cm 0.cm 0.cm},clip]{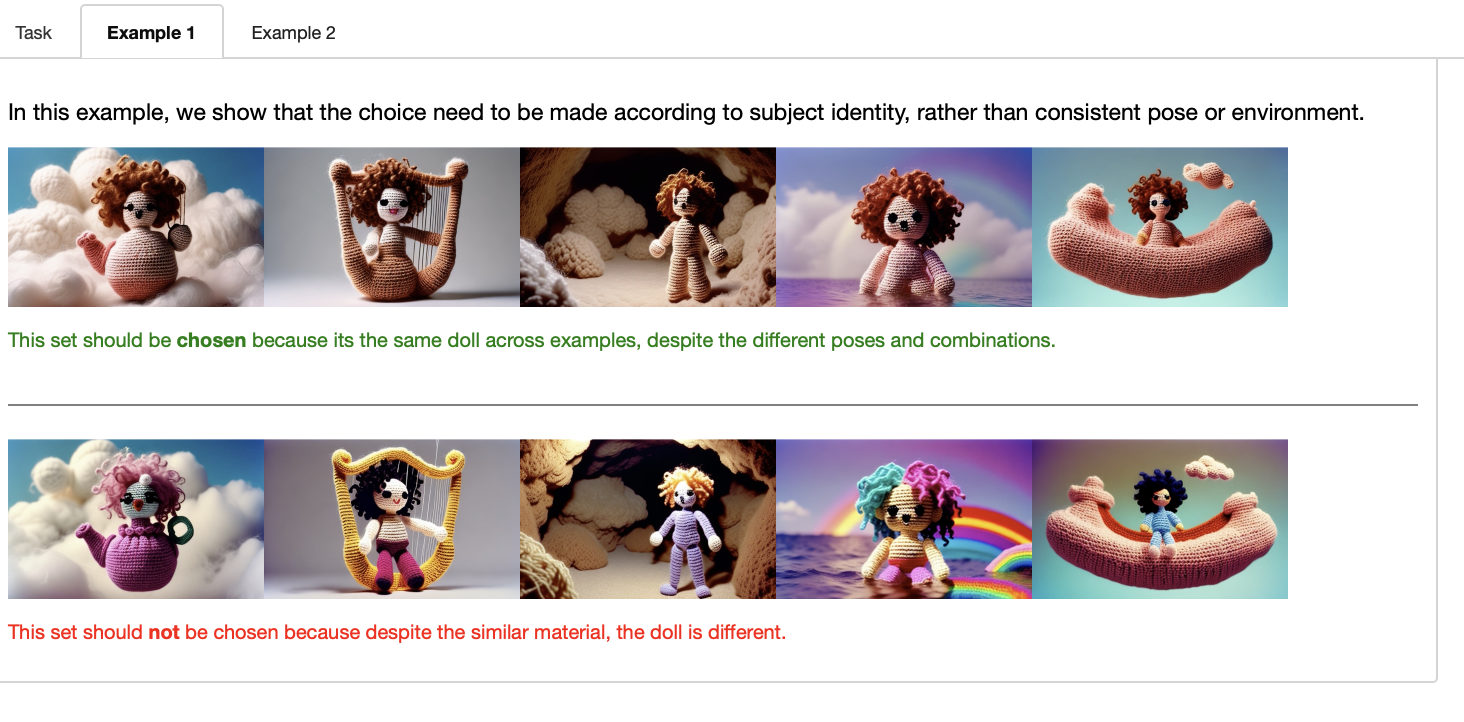} %
\includegraphics[width=0.48\textwidth, trim={0.cm 0.cm 0.cm 0.cm},clip]{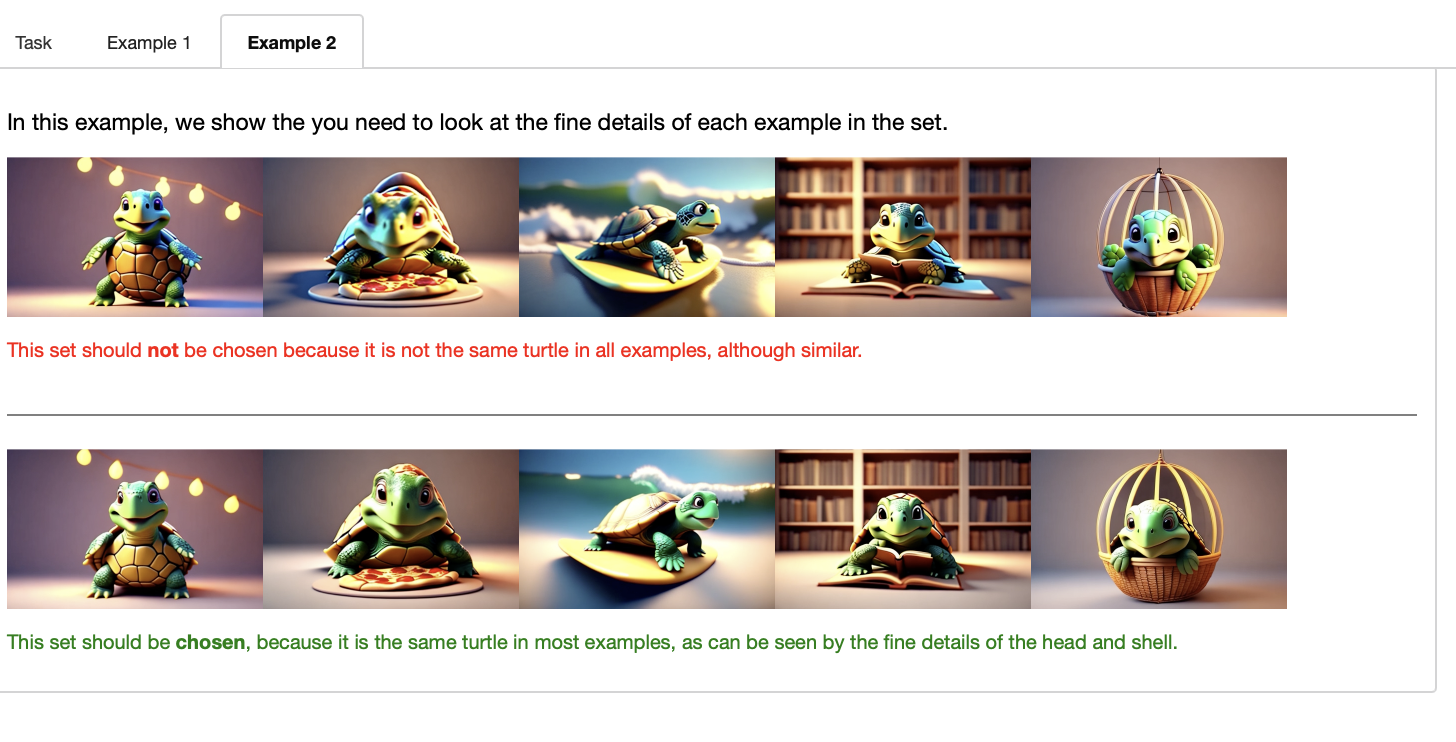} %
\caption{Examples provided in the user study for visual set consistency. }
\label{fig_amt1_examples}
\end{flushleft}
\end{figure}

\begin{figure}[ht]
\includegraphics[width=\textwidth, trim={0.cm 0.cm 0.cm 0.cm},clip]{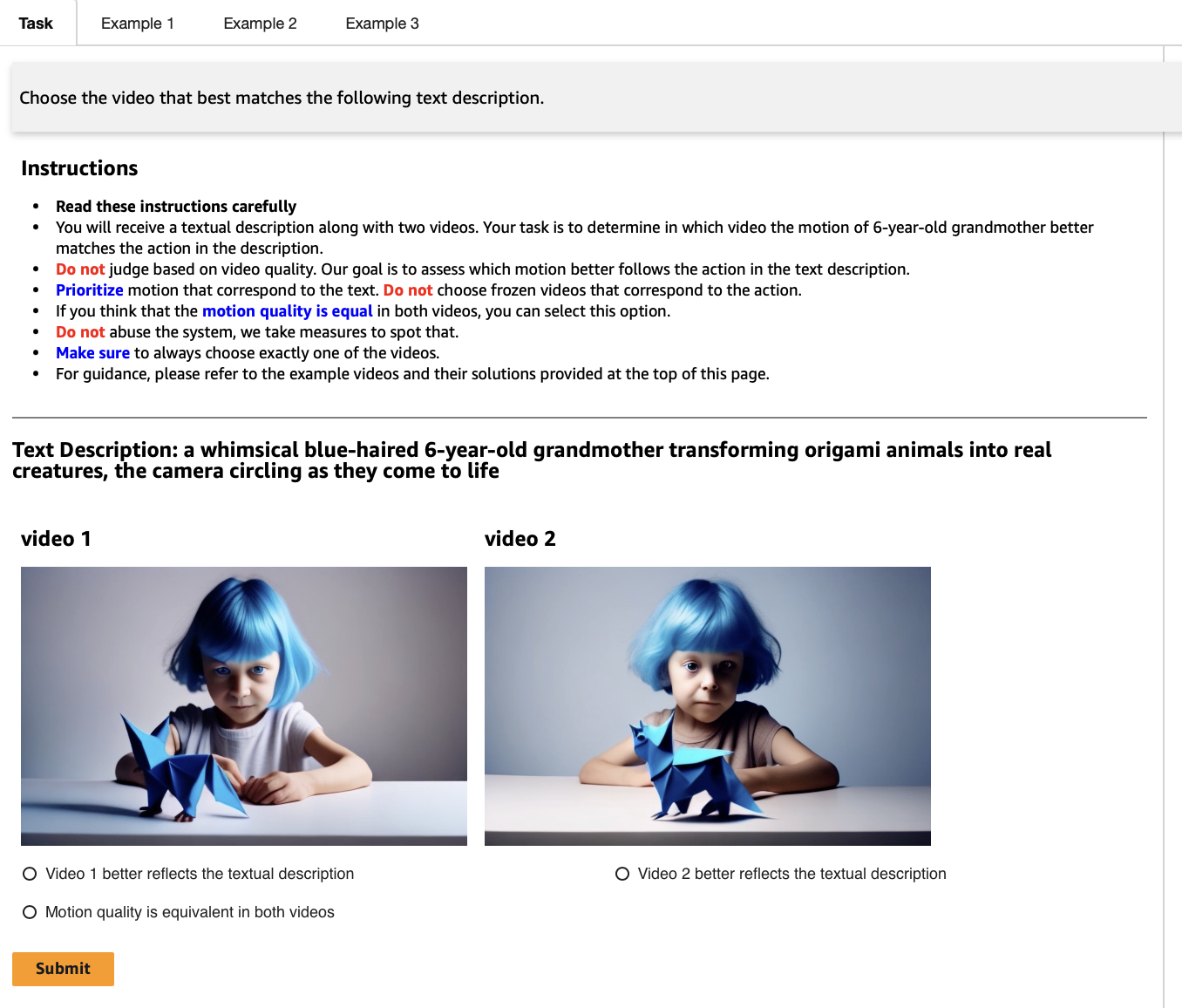} %
\vspace{-15pt}
\caption{One trial of the text-motion alignment user study. }
\label{fig_amt2_task}

\begin{flushleft}
\includegraphics[width=0.48\textwidth, trim={0.cm 0.cm 0.cm 0.cm},clip]{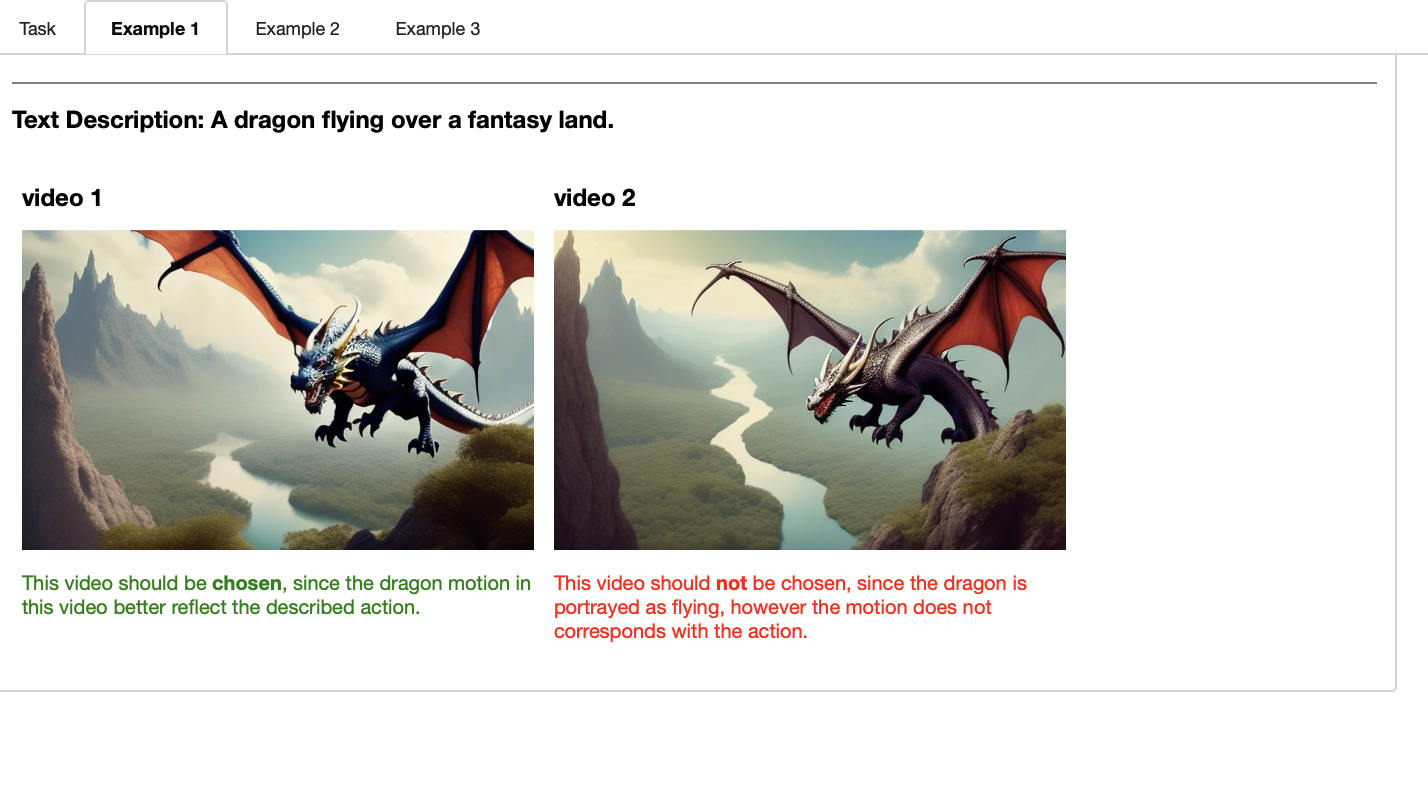} %
\includegraphics[width=0.48\textwidth, trim={0.cm 0.cm 0.cm 0.cm},clip]{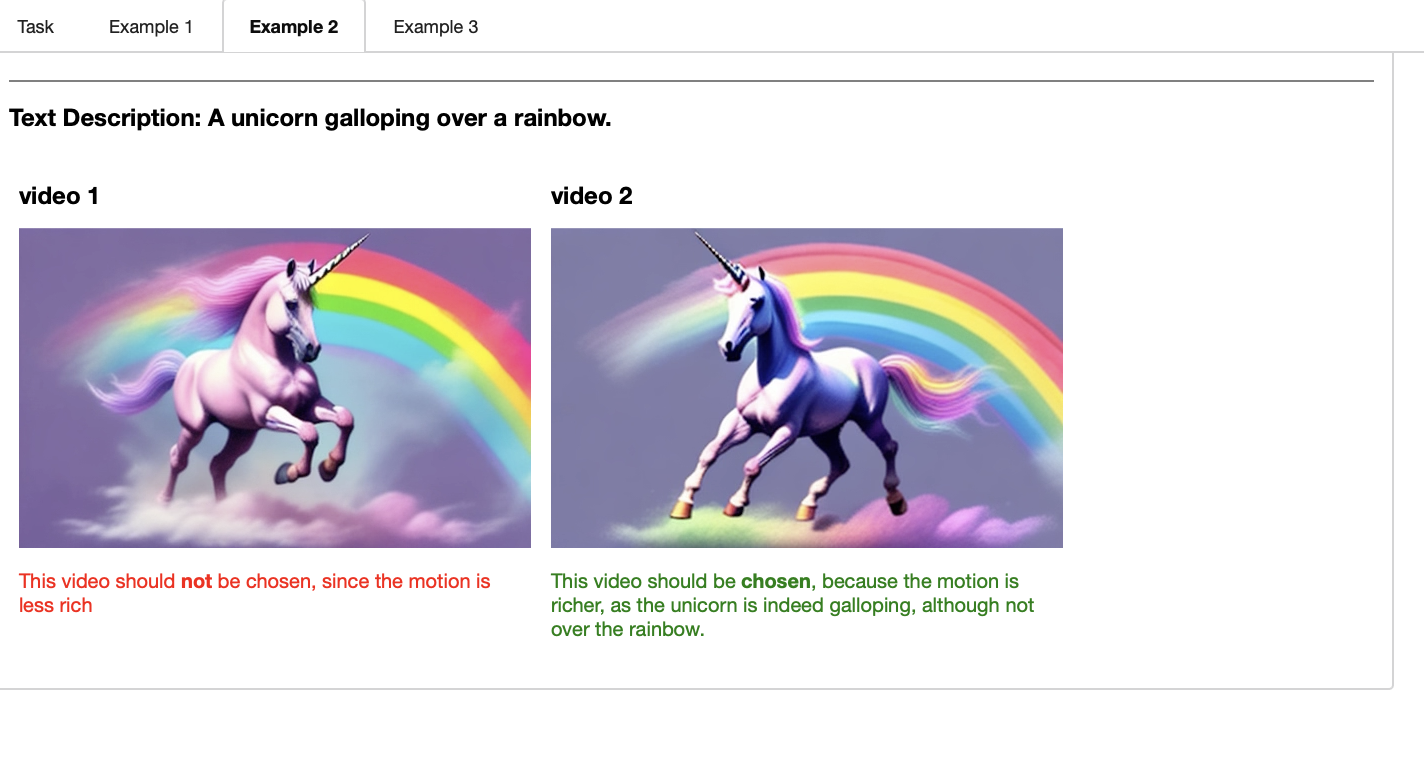} %
\includegraphics[width=0.48\textwidth, trim={0.cm 0.cm 0.cm 0.cm},clip]{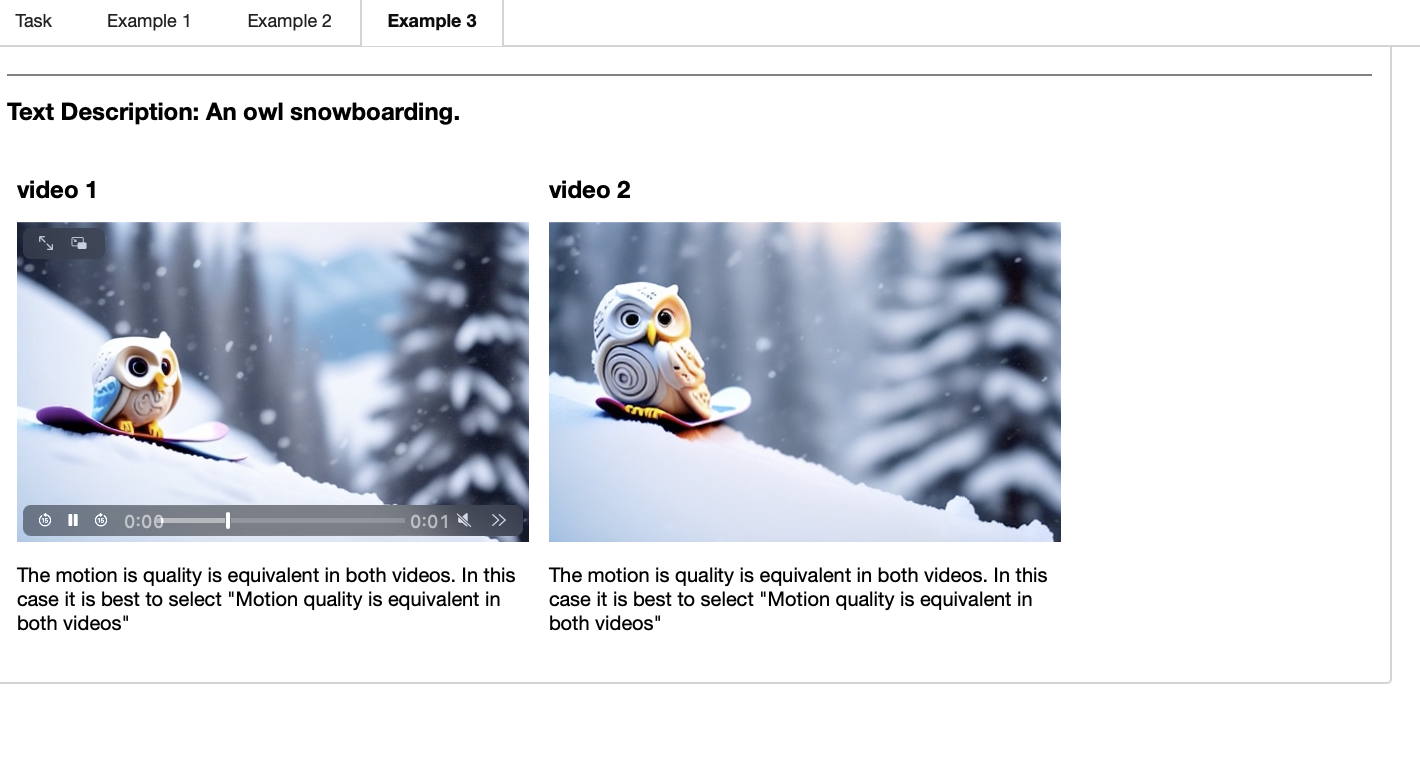} %
\vspace{-10pt}
\caption{Examples provided in the user study for text-motion alignment. } 
\label{fig_amt2_examples}
\end{flushleft}
\end{figure}

\end{document}